\documentclass[journal,10pt,twocolumn]{IEEEtran}

\date{}
\usepackage{graphicx}
\usepackage{epsfig}
\usepackage{subfigure}
\usepackage{amssymb}
\usepackage{amsbsy}
\usepackage{amsmath}
\usepackage{steinmetz}
\usepackage{hyperref}
\usepackage{stmaryrd}
\usepackage{cite}
\usepackage{url}
\usepackage{amsfonts}

\DeclareMathOperator*{\argmin}{\arg\min}  
  
\usepackage{caption}
\usepackage{color}
\usepackage{float}
\usepackage{times,amsmath,epsfig}
\usepackage{xspace,latexsym,syntonly}
\usepackage{amssymb}
\usepackage{textcomp}
\usepackage{xcolor}

\newtheorem{definition}{Definition}
\newtheorem{theorem}{Theorem}

\newtheorem{lemma}{Lemma}

\newcommand{\qed}{\hfill\blacksquare}

\usepackage{algorithm}
\usepackage{algpseudocode}

\begin{document}
\title{Accelerated Gradient Descent Learning over Multiple Access Fading Channels}
\author{Raz Paul, Yuval Friedman, Kobi Cohen
\thanks{$\copyright$ 2021 IEEE. Personal use of this material is permitted. Permission from IEEE must be obtained for all other uses, in any current or future media, including reprinting/republishing this material for advertising or promotional purposes, creating new collective works, for resale or redistribution to servers or lists, or reuse of any copyrighted component of this work in other works.}
\thanks{Raz Paul, Yuval Friedman and Kobi Cohen are with the School of Electrical and Computer Engineering, Ben-Gurion University of the Negev, Beer Sheva 8410501 Israel. Email: razpa@post.bgu.ac.il, yufri@post.bgu.ac.il, yakovsec@bgu.ac.il}
\thanks{This research was supported by the ISRAEL SCIENCE FOUNDATION (grant No. 2640/20).}
}

\maketitle

\begin{abstract}
We consider a distributed learning problem in a wireless network, consisting of $N$ distributed edge devices and a parameter server (PS). The objective function is a sum of the edge devices' local loss functions, who aim to train a shared model by communicating with the PS over multiple access channels (MAC). This problem has attracted a growing interest in distributed sensing systems, and more recently in federated learning, known as over-the-air computation. In this paper, we develop a novel Accelerated Gradient-descent Multiple Access (AGMA) algorithm that uses momentum-based gradient signals over noisy fading MAC to improve the convergence rate as compared to existing methods. Furthermore, AGMA does not require power control or beamforming to cancel the fading effect, which simplifies the implementation complexity. We analyze AGMA theoretically, and establish a finite-sample bound of the error for both convex and strongly convex loss functions with Lipschitz gradient. For the strongly convex case, we show that AGMA approaches the best-known linear convergence rate as the network increases. For the convex case, we show that AGMA significantly improves the sub-linear convergence rate as compared to existing methods. Finally, we present simulation results using real datasets that demonstrate better performance by AGMA.
\end{abstract}


\section{Introduction}

We consider a distributed learning problem in a wireless network, consisting of $N$ distributed edge devices (i.e., nodes) and a parameter server (PS). The objective function is a sum of the nodes' local loss functions, who aim to train a shared model by communicating with the PS over multiple access channels (MAC). Specifically, the PS aims at solving the following  optimization problem: 
\begin{equation}
\label{eq:theta_star_intro}
\boldsymbol{\theta^*} = \argmin_{\boldsymbol{\theta} \in \Theta} \frac{1}{N}\sum_{n=1}^N f_n(\boldsymbol{\theta})
\end{equation}
based on data received from the nodes. The model $\boldsymbol{\theta}\in\Theta\subset\mathbb{R}^d$ is a $d\times 1$ parameter vector which needs to be optimized. The solution $\boldsymbol{\theta^*}$ is known as the empirical risk minimizer. In machine learning (ML) tasks, the loss function is typically given by 

$f_n(\boldsymbol{\theta})=\ell(\boldsymbol{\theta};\boldsymbol{x}_n, y_n)$,
which is the loss given the pair sample $(\boldsymbol{x}_n, y_n)$ (e.g., $\boldsymbol{x}_n$ refers to the input vector and $y_n$ is the corresponding label) with respect to the model parameter $\boldsymbol{\theta}$. The goal is to train the algorithm so as to find a shared model $\boldsymbol{\theta}$ that transforms the input vector $\boldsymbol{x}$ into the desired output $y$.

Traditional ML algorithms solve (\ref{eq:theta_star_intro}) in a centralized manner. This approach requires to store all data at the PS, which in turn implements a centralized optimizer (e.g., gradient descent (GD)-type algorithm). However, the increasing demand of mobile applications, such as 5G, IoT, and cognitive radio applications, makes centralized ML algorithms inefficient in terms of communication resources required to upload the entire distributed raw data to the PS \cite{chen2020joint}. Furthermore, due to privacy concerns, local data should be stored at the local edge devices \cite{mcmahan2017communication}. Therefore, in recent years federated learning was suggested to solve these issues and consequently received a growing attention. In federated learning, the training is distributed among a large number of nodes, were each node operates local processing based on its local data, and transmits an output (e.g., local gradient) to the PS. The PS then aggregates the received data from the nodes to update the global model, transmits the updated model back to the nodes, and so on. The problem  finds applications in distributed sensing and control systems as well (see related  work  in  Section \ref{ssec:learning}). 

\subsection{Related work} 
\label{ssec:learning}

In traditional inference and learning algorithms in communication networks, data is transmitted over orthogonal channels (e.g., TDM, FDM), which increases the bandwidth requirement linearly with the number of nodes $N$, as well as increases the power consumption due to the additive noise in each dimension. By contrast, learning methods over MAC, known as \emph{over-the-air computation} \cite{amiri2020machine}, exploit the inherent nature of the wireless channel to make the computation over the air (e.g., by summing analog signals required to update the trained model). As a result, the PS receives a superposition of the transmitted signals, which yields a (variation of a) sufficient statistics for the learning task. The number of dimensions used for transmitting the data over MAC is thus independent of $N$, which results in high power and bandwidth efficiency. 

In this paper we focus on gradient-based learning over MAC, in which each node transmits a local analog gradient-type signal of the current model update. The PS receives an aggregated signal which represents a global noisy (due to the receiver's additive noise) distorted (due to fading channel effect) gradient-type signal which is used to update the model. Learning algorithms based on similar gradient-based methods have been studied in recent years (see \cite{ amiri2019collaborative, zhu2019broadband,  zeng2019energy,  amiri2019over, sery2020analog,  abdi2020analog, amiri2020machine, chang2020communication,  amiri2020federated, ozfatura2020distributed} and references therein). In \cite{amiri2020machine, amiri2019over, amiri2020federated, ozfatura2020distributed}, the authors developed the compressed analog distributed stochastic gradient descent (SGD) method, in which a sparse parameter gradient vector is transmitted by the nodes over MAC. In \cite{amiri2020federated}, power control is used to eliminate the fading distortion, where nodes in deep fading do not transmit to satisfy the power constraint. In \cite{amiri2019collaborative}, the fading distortion is mitigated at the receiver by using multiple antennas, where the fading diminishes as the number of antennas approaches infinity. Channel communication characteristics have been further studied in \cite{abdi2020analog}. In our previous work \cite{sery2020analog}, we have developed and analyzed gradient-based learning without using power control or beamforming to cancel the fading effect. In \cite{zhu2018low, zhu2019broadband}, the authors developed the federated edge learning algorithm that schedules entries of the  gradient vector based on the channel condition. Energy-efficiency aspects have been studied in \cite{zeng2019energy}. Quantization methods of gradient transmissions were developed in \cite{chang2020communication}. Other aspects of learning over MAC that were studied recently are over-the-air federated learning with heterogeneous \cite{abad2019hierarchical,ahn2019wireless, sery2020over, gafni2021federated} or redundant \cite{sun2019energyaware} data, over-the-air aggregation techniques via MAC with known channel states \cite{ahn2019wireless,sun2019energyaware}, over-the-air computation with sub-Gaussian fading and noise distributions \cite{frey2020over}, digital gradient transmissions \cite{chang2020communication}, and privacy over MAC \cite{seif2020wireless}.

In earlier years, distributed inference and learning has been widely studied under model-dependent settings, where the observation distributions are assumed to be known. Traditional communication methods transmit data signals over orthogonal channels. Various methods that reduce the number of transmissions by scheduling nodes with better informative observations were developed in \cite{Patwari_Hierarchical_2003, Appadwedula_Decentralized_2007, Blum_Energy_2008, blum2011ordering, braca2011asymptotically, Cohen_Energy_2011,  braca2012single, cohen2015active, zhang2017ordering, sriranga2018energy, huang2019active}. Another recent method is the Lazily Aggregated Gradient (LAG) algorithm \cite{chen2018lag} that executes GD-type iterates in which every node computes and transmits the difference between the gradient at each iteration and the gradient at the previous update. However, the bandwidth increases linearly with the number of nodes $N$ when using schemes that transmit on orthogonal channels (i.e., dimension per node). By contrast, as explained above, the bandwidth requirement by over-the-air learning methods that use MAC to aggregate transmitted signals does not increase with $N$. Past research focused on transmission schemes over MAC where the observation distributions are assumed to be known (see \cite{mergen2006type, Mergen_Asymptotic_2007, Liu_Type_2007, Marano_Likelihood_2007, zhang2016event, anandkumar2007type, li2011decision,  maya2015optimal, maya2015exploiting} and our previous work \cite{cohen2013performance,  cohen2019time, cohen2018spectrum}). However, all these studies assumed that the observation distributions are known to the nodes or to the network edge, which are assumed to be unknown in this paper inspired by distributed machine learning and federated learning applications. 

\subsection{Main Results}
\label{Intro:main}

In this paper we focus on accelerating the convergence of gradient-based learning over noisy fading MAC. The motivation is inspired by the fact that acceleration can be made in a centralized noiseless distortion-free setting. However, this question remained open in previous studies of over-the-air gradient-based learning, which we aim to solve. Specifically, our contributions are summarized below. 

First, we develop a novel Accelerated Gradient-descent Multiple Access (AGMA) algorithm to achieve our goal. By contrast to existing over-the-air gradient-based learning methods that compute the gradient directly with respect to the last update, in AGMA, each node computes a momentum-based gradient that uses the last two updated models. AGMA is advantageous in terms of practical implementations, since it does not use power control or beamforming to cancel the channel gain effect as in \cite{zhu2018low, amiri2019collaborative, zhu2019broadband,  amiri2019over,   abdi2020analog, amiri2020machine, amiri2020federated, ozfatura2020distributed}. It should be noted that schemes that correct the channel gains (for instance, by dividing the gradient signal at the transmitters by the channel gain to avoid distortion at the receiver) might still suffer from channel estimation errors. Also, in schemes that censor transmissions by nodes, depending on their experienced channel gains (to satisfy a transmission power constraint or limit the dynamic range of the transmitted signal), the global received gradient aggregates local gradients which are multiplied by one (for transmitted signals that correct the channel gains) and zero (for censored signals). As a result, these models generate a global noisy distorted gradient. Thus, the analysis in this paper contributes to address these models as well. The PS updates the model based on the noisy distorted momentum-based gradient directly. This type of transmission scheme that uses noisy distorted signals over MAC for inference tasks was analyzed under various settings (see \cite{mergen2006type, Mergen_Asymptotic_2007, anandkumar2007type, cohen2013performance, cohen2018spectrum, sery2020analog} and references therein, as well as related work in Section \ref{ssec:learning}). In this paper, we first develop and analyze this type of transmission scheme in the setting of momentum-based gradient learning over noisy fading MAC. Note that standard SGD with momentum was not analyzed under noisy fading MAC in previous studies. The design in this paper is in a family of over-the-air learning algorithms, which have attracted a growing interest in recent years. Thus, the implementation via analog signal transmissions of the accelerated SGD with momentum is fundamentally different than classic centralized/noiseless implementations. Specifically, the transmissions require $d$ orthogonal waveforms (one for each gradient dimension), by contrast to direct access of $N\cdot d$ data signals in classic implementations (which scales with the number of nodes $N$). Second, the algorithm design requires careful energy scaling laws for signal transmissions to guarantee convergence, which is absent in classic centralized/noiseless implementations. Also, our design does not require power control to vanish the distortion due to the channel fading in the receiver. These effects are taken into account by the design of the energy scaling laws of signal transmissions and reflected in the error analysis as well.

Second, we analyze AGMA theoretically and establish a finite-sample bound of the error for both convex and strongly convex loss functions with Lipschitz gradient. We develop specific design principles for the learning step and power scaling laws for signal transmissions to guarantee convergence under momentum-based gradient distortion due to the fading effect and additive noise at the receiver. For the strongly convex case, we show that AGMA approaches the best-known linear convergence rate $O(c^k)$ as $N\rightarrow\infty$, where $k$ is the number of iterations and $0<c<1$ is a finite constant.
In this case, the additive noise affects the error via the following term: $C\frac{d\sigma_{w}^{2}}{E_{N}N^{2}}$, where $C=\sqrt{\frac{\widetilde{L_\beta}}{\mu}}\frac{\beta}{\mu_{h}}$ is a constant depending on the system model (where $\widetilde{L_\beta}$ is an auxiliary constant which depends on the Lipschitz continuity as defined in \eqref{th:L_beta}, $\mu$ is the strong convexity constant, $\beta$ is the stepsize, and $\mu_h$ is the channel gain mean). As a result, we can set the transmission power to $E_N=\Omega\left(N^{\epsilon -2}\right)$, for some $\epsilon>0$, to eliminate this term as $N$ increases\footnote{Note that $E_N=\Omega\left(N^{\epsilon -2}\right)$ refers to Big Omega notation in complexity theory (Knuth), i.e., there exist $k>0, N_0$ such that for all $N>N_0$, we have: $E_N\geq k N^{\epsilon -2}$.}.

For the convex case, we show that AGMA improves the sub-linear convergence rate from $1/k$ to $1/k^2$ as compared to existing methods as $N\rightarrow\infty$. In this case, the additive noise affects the error via the following term: $\frac{\beta}{\mu_{h}}\cdot\frac{d\sigma_{w}^{2}}{E_{N}N^{1+\epsilon}}$, for $0<\epsilon<1$, and iterations $k<\lfloor N^{1-\epsilon}\rfloor$. As a result, we can set the transmission power to $E_N=\Omega\left(N^{-1-\epsilon'}\right)$, for some $0<\epsilon'<\epsilon$, to eliminate this term as $N$ increases. We infer from these results that we can improve the learning accuracy by increasing the number of nodes used in the learning task, while the total invested transmission power in the network can be set arbitrarily close to zero.

It is worth noting that the effect of diverged error in accelerated GD algorithms is well known when handling noisy gradients, and heuristic restarted methods that avoid momentum are often adopted when $k$ is large (see e.g., \cite{wang2020scheduled},\cite{Adaptive_Restart} and references therein). To the best of our knowledge, our theoretical analysis provides the first results that guarantee convergence analytically in accelerated GD over noisy fading MAC. To tackle this challenge, we establish a new stochastic version of auxiliary sequence method, which considers both gradient distortion and additive noise. By contrast to existing methods that fail to bound the error when handling noisy gradients, we exploit the structure of MAC transmissions to control the error by the fact that the global gradient is computed distributedly, where the distortion and noise effects can be mitigated by controlling the network size and the transmission power. Based on this observation, we design the new auxiliary function such that it is computed based on the noisy estimate of the gradient $\boldsymbol{v}_{k}$ over the fading MAC. We introduced a new trick to correct the auxiliary function by a guarding term $\epsilon_N$ (which decreases with $N$) used to guard against the diverged error. Under mild conditions on the stochastic processes, we are able to upper bound the error by iterating over the estimate updates. This development resolves the strongly convex case for all $k$ by using an auxiliary control sequence which can be bounded by a condition number (defined in 
\eqref{eq:condition_number} in the analysis). By contrast, in the convex case, the condition number cannot be properly defined. Therefore, to tackle this challenge we exploit the structure of MAC transmissions to bound the error. Since the global noisy distorted gradient is a sum of $N$ local noisy corrupted gradients, computed distributedly, we can bound the auxiliary control sequence by induction over the number of iterations, which depends on $N$. Moreover, by contrast to heuristic restarted methods which are often adopted to avoid momentum when using noisy gradient when $k$ is large (to avoid diverged error) (see e.g., \cite{wang2020scheduled},\cite{Adaptive_Restart} and references therein), our results provide the first analytic solution in restarted methods with theoretical guarantees, depending on the iteration number and the network size.

Third, we examine the performance of AGMA numerically in three different settings using real datasets. In the first setting, we simulate a federated learning task used to predict a release year of a song from its audio features. We use the popular Million Song Dataset \cite{Bertin-Mahieux2011} for this task. In the second setting, we simulate a distributed learning task for estimation in radar systems. We use the popular Ionosphere dataset collected by a radar system in Goose Bay, Labrador available by UCI Machine Learning Repository \cite{Dua:2019} for this task. In the third setting, we trained a neural network (NN) for handwritten digit classification using the MNIST dataset \cite{mnist}, where the theoretical conditions for the analysis are not met. The simulation results demonstrate very good performance of AGMA as compared to existing methods in all three experiments.

\section{Preliminaries}

We start by providing a background knowledge used in the optimization and learning literature that will be used throughout the paper (for more details on the background provided in this section the reader is referred to \cite{Nesterov2004IntroductoryLO}). Below, we define a function $f(\cdot)$ with $L$-Lipschitz continuous gradient $\boldsymbol{\nabla} f(\cdot)$.

\noindent
\begin{definition}
A function $f(\boldsymbol{x})$ with domain $X$ has a Lipschitz continuous gradient if it is continuously differentiable for any $\boldsymbol{x}\in X$, and the inequality
\begin{equation}\label{L_lemma}
\displaystyle ||\boldsymbol{\nabla} f(\boldsymbol{x}) - \boldsymbol{\nabla} f(\boldsymbol{y})|| \leq L||\boldsymbol{x}-\boldsymbol{y}||
\end{equation}
holds for all $\boldsymbol{x},\boldsymbol{y}\in X$. The constant $L$ is called the Lipschitz constant.\vspace{0.1cm}
\end{definition}

Next, we define the strong convexity property of a function $f(\cdot)$.

\noindent
\begin{definition}
A function $f(\boldsymbol{x})$ with domain $X$ is $\mu$-strongly convex if it is continuously differentiable for any $\boldsymbol{x} \in X$ and the inequality
\begin{equation}\label{mu_lemma}
\displaystyle \left<\boldsymbol{\nabla} f(\boldsymbol{x}) - \boldsymbol{\nabla} f(\boldsymbol{y}), \boldsymbol{x}-\boldsymbol{y} \right> \geq \mu||\boldsymbol{x}-\boldsymbol{y}||^2
\end{equation}
holds for all $\boldsymbol{x},\boldsymbol{y}\in X$. The constant $\mu$ is called the strong convexity constant.
\end{definition}

Finally, we present useful lemmas of the linearity of strong convexity and Lipschitz continuous properties that will be used in the analysis in this paper.

\noindent
\begin{lemma}
\label{lem:linear_Lipschitz}
Consider two Lipschitz continuous functions, $f(\boldsymbol{x}), g(\boldsymbol{x})$ with Lipschitz constants $L^f$ and $L^{g}$, 
respectively. Then, the function $\psi(\boldsymbol{x})= \alpha f(\boldsymbol{x}) +\beta g(\boldsymbol{x})$ is Lipschitz continuous with Lipschitz constant $L^\psi =\alpha L^{f}+ \beta L^{g}$.
\end{lemma}

\noindent
\begin{lemma}
\label{lem:linear_strong_convex}
Consider two strongly convex functions, $f(\boldsymbol{x}), g(\boldsymbol{x})$ with constants $\mu^f$ and $\mu^{g}$, respectively. Then, the function $\psi(\boldsymbol{x})= \alpha f(\boldsymbol{x}) +\beta g(\boldsymbol{x})$ is strongly convex with constant  $\mu^\psi =\alpha \mu^{f}+\beta \mu^{g}$.
\end{lemma}\vspace{0.1cm}

The proofs for the lemmas in this section can be found in \cite{Nesterov2004IntroductoryLO}.

\section{Network Model and Problem Statement}
\label{sec:system}

\begin{figure}[htbp]
\begin{center}{\scalebox{0.25}
  {\epsfig{file=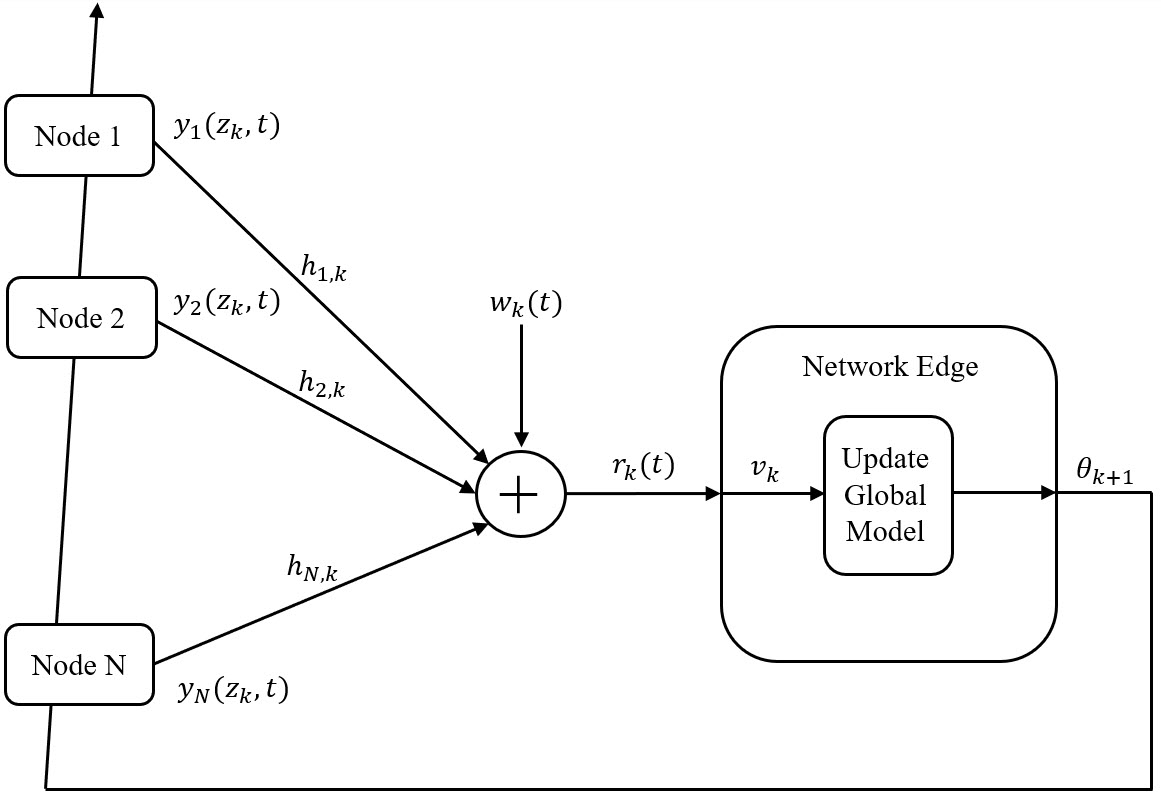}} }
  \caption{An illustration of the wireless network considered in this paper. Each node communicates directly with the PS at the network edge, who updates the global model and broadcasts the updated model back to the nodes.}
  \label{fig:scheme}
  \end{center}
  \end{figure}

We consider a wireless network consisting of $N$ nodes (i.e., edge devices) indexed by the set $\mathcal{N} = \{1,2,...,N\}$ and a PS at the network edge. Each node communicates directly with the PS. The transmission time is slotted, and indexed by $t_0, t_1, t_2, ...$. Each node $n$ experiences at time $t_k$ a block fading channel $\tilde{h}_{n,k}$ with gain $h_{n,k}\triangleq|\tilde{h}_{n,k}|\in \mathbb{R}_+$ and phase $\phi_{n,k}\triangleq\phase{\tilde{h}_{n,k}}\in \left\{x\in\mathbb{R}|-\pi\leq x\leq\pi\right\}$. The channel fading is assumed to be i.i.d. across time and nodes, with mean $\mu_h$ and variance $\sigma_h^2$ as in \cite{zhu2019broadband, amiri2019over, sery2020analog, amiri2020federated}. An illustration of the network is presented in Fig. \ref{fig:scheme}. 

Each node is associated with a local loss function $f_n$, and the objective function is a sum of the nodes' local loss functions:
\begin{equation}
\label{eq: F definition}
F(\boldsymbol{\theta})\triangleq\frac{1}{N} \sum_{n=1}^N f_n(\boldsymbol{\boldsymbol{\theta}}).
\end{equation} 
The objective of the PS is to solve the following optimization problem:
\begin{equation}
\label{eq:theta_star}
\boldsymbol{\theta^*} = \argmin_{\boldsymbol{\theta} \in \Theta} \;F(\boldsymbol{\theta})
\end{equation}
based on data received from the nodes. As commonly assumed when analyzing GD-based methods (see e.g., \cite{Nemirovski2009}, \cite{Shalev2011} and subsequent studies), for purposes of analysis we assume that $f_n$ is convex (and also strongly convex in Section \ref{ssec:strongly_lip}), has Lipschitz gradient with Lipschitz constant $L_n$, and a bounded expected local gradient power: $\mathbb{E}[\|\boldsymbol{\nabla{f_{n}(\cdot)}}\|^{2}] \leq G$. The term $\boldsymbol{\theta}\in\Theta\subset\mathbb{R}^d$ is the $d\times 1$ parameter vector which needs to be optimized. The solution $\boldsymbol{\theta^*}$ is known as the empirical risk minimizer. Each node $n$ is aware only of its local loss function $f_n$. As commonly assumed in the literature on GD-based optimization methods, it is assumed that $\boldsymbol{\nabla{f_n}}(\boldsymbol{\theta})$ exists for all $n\in\mathcal{N}$ and $\boldsymbol{\theta} \in \Theta$, and is computable. Otherwise, subgradient methods should be applied\cite{Nesterov2004IntroductoryLO}. It is worth noting that throughout the paper we are interested in analyzing an objective function $F(\boldsymbol{\theta})$ which is the average over local functions 
$f_n(\boldsymbol{\theta})$ (see \eqref{eq: F definition}). We will analyze both convex and strongly-convex cases. Therefore, by Lemmas \ref{lem:linear_Lipschitz}, \ref{lem:linear_strong_convex}, if we assume $\mu_n$-strong convexity of $f_n(\boldsymbol{\theta})$, this implies $\left(\frac{1}{N}\sum_{n=1}^N \mu_{n}\right)$-strong convexity of $F(\boldsymbol{\theta})$. Similarly, if we assume $L_n$-Lipschitz gradient of $f_n(\boldsymbol{\theta})$, this implies   $\left(\frac{1}{N}\sum_{n=1}^N L_{n}\right)$-Lipschitz gradient of $F(\boldsymbol{\theta})$.

\section{The Proposed Accelerated Gradient-descent Multiple Access (AGMA) Algorithm}
\label{sec:AGMA}

In this section we present the Accelerated Gradient-descent Multiple Access (AGMA) algorithm to solve the objective (\ref{eq:theta_star}). As commonly implemented by over-the-air gradient aggregation methods for learning over MAC \cite{zhu2018low, amiri2019collaborative, zhu2019broadband,  zeng2019energy,  amiri2019over, sery2020analog,  abdi2020analog, amiri2020machine, chang2020communication,  amiri2020federated, ozfatura2020distributed}, all nodes transmit a function of the local gradient to the PS simultaneously using common analog waveforms. The PS updates the estimate based on the received data and broadcasts the updated estimate back to the nodes, and so on until convergence. An illustration is given in Fig. 
\ref{fig:scheme}. A key difference in AGMA is that by contrast to existing methods that compute the gradient with respect to the last update directly, in AGMA, each node computes a momentum-based gradient. Furthermore, AGMA does not use power control or beamforming to cancel the channel gain effect as in \cite{zhu2018low, amiri2019collaborative, zhu2019broadband,  amiri2019over,   abdi2020analog, amiri2020machine, amiri2020federated, ozfatura2020distributed}. The PS updates the model based on the noisy distorted momentum-based gradient directly. We will show theoretically in Section \ref{sec:performance} that for the strongly convex case, AGMA approaches the best-known linear convergence rate as the network size increases. For the convex case, we will show that AGMA improves the sub-linear convergence rate as compared to existing methods. In Section \ref{sec:sim} we present simulation results that demonstrate better performance by AGMA.

We next describe AGMA in detail. As commonly assumed in the over-the-air learning literature, we assume that the channel state information (CSI) is available for each node before transmitting the data signal to the PS, which is typically done in communication networks by estimating the channel state from a pilot signal broadcast by the network edge \cite{Mergen_Asymptotic_2007, cohen2013performance, wimalajeewa2015wireless, yu2020optimizing}. Let $\boldsymbol{\theta_k}$ be the updated estimate of the parameter at iteration $k$. All nodes and the PS compute a momentum-based updated model $\boldsymbol{z_{k}} $ based on the two-step history by: 
\begin{equation}
\boldsymbol{z_{k}} = \boldsymbol{\theta_{k}} + \eta_{k-1}(\boldsymbol{\theta_{k}}-\boldsymbol{\theta_{k-1}}).
\label{eq:z_k}
\end{equation}
The PS stores $\boldsymbol{z_{k}}$ and uses it later to update the model. At the nodes side, each node computes its local gradient $\boldsymbol{\nabla{f_n}}(\boldsymbol{z_k})$ with respect to $\boldsymbol{z_{k}}$. The algorithm is initialized by initial model $\boldsymbol{\theta_{0}}$, and $\boldsymbol{z_{0}}$ is initialized by $\boldsymbol{z_{0}}=\boldsymbol{\theta_{0}}$. The term $\eta_k$ is the momentum coefficient, and it will be designed in Section \ref{sec:performance}. Then, all nodes transmit simultaneously the following analog data signal:
\begin{equation}
y_{n}(\boldsymbol{z_{k}},t)=\sqrt{E_N}e^{-j\phi_{n, k}}  \boldsymbol{\nabla{f_n}}(\boldsymbol{z_k})^T \boldsymbol{s(t)}, 
\label{eq:y_n_z_t}
\end{equation}
where $E_N$ is a transmission power-coefficient set to satisfy the power requirement, $e^{-j\phi_{n, k}}$ is used to correct the phase reflection to yield coherent aggregated signals at the receiver, as suggested in past studies (e.g., \cite{Mergen_Asymptotic_2007, cohen2013performance, wimalajeewa2015wireless, sery2020analog,  yu2020optimizing, sery2020over}), and $\boldsymbol{s(t)}=\left[s_1(t), ..., s_d(t)\right]^T$ is a column vector of $d$ orthogonal baseband equivalent normalized waveforms, as suggested in \cite{sery2020analog}. 
Also, the fact that AGMA does not use power control or beamforming to correct the channel gain as in \cite{zhu2018low, amiri2019collaborative, zhu2019broadband,  amiri2019over,   abdi2020analog, amiri2020machine, amiri2020federated, ozfatura2020distributed} is advantageous in various network applications. For example, we can set the average transmission power of the data signal based on the observations statistics solely independently of the channel statistics. Furthermore, it avoids increasing the dynamic range of the transmitted signal due to the channel gain variance, which simplifies the hardware implementations. Finally, note that schemes that correct the channel gains might still suffer from channel estimation errors which cause signal distortions. Thus, the analysis in this paper captures this setting as well.

The received signal at the PS is given by a superposition of all transmitted signals:
\begin{equation}
\begin{array}{c}
 \displaystyle r_k(t) = \sum_{n=1}^{N}\sqrt{E_N}  h_{n,k} \boldsymbol{\nabla{f_n}}(\boldsymbol{z_k})^T \boldsymbol{s(t)}+ w_k(t),\vspace{0.2cm}\\\hspace{5cm}
 \displaystyle
 t_k\leq t <t_k+T,
\end{array}
\label{eq:r_k_t}
\end{equation}
where $w_k(t)$ is a zero-mean additive Gaussian noise process at time $t_k\leq t <t_k+T$. After matched filtering $r_k(t)$ and averaging, we have:
\begin{equation} 
\boldsymbol{v_k}\triangleq \frac{1}{N} \sum_{n=1}^{N}  h_{n,k} \boldsymbol{\nabla{f_n}}(\boldsymbol{z_k}) + \boldsymbol{w_k},
\label{eq:v_k_definition}
\end{equation}
where $\boldsymbol{w_k}\sim \mathcal{N}(0,\frac{\sigma_w^2}{N^2E_N} \boldsymbol{I_d})$, and $\boldsymbol{I_d}$ is the $d\times d$ identity matrix. Then, based on the stored $\boldsymbol{z_k}$ in (\ref{eq:z_k}), the PS updates the model $\boldsymbol{\theta_{k+1}}$ by:
\begin{equation} 
\boldsymbol{\theta_{k+1}} = \boldsymbol{z_k} - \beta \boldsymbol{v_k}, 
\label{eq:theta_k_1}
\end{equation}
and broadcasts the updated model back to the nodes via error-free channel\footnote{Note that we can use digital communication schemes to broadcast the estimate back to the nodes, where the bandwidth requirement does not scale with $N$. Thus, we assume an error-free channel in this phase, as commonly assumed in the over-the-air learning literature\cite{zhu2018low, amiri2019collaborative, zhu2019broadband,  zeng2019energy,  amiri2019over, sery2020analog,  abdi2020analog, amiri2020machine, chang2020communication,  amiri2020federated, ozfatura2020distributed}.}, which ends iteration $k$. In iteration $k+1$, the nodes and the PS compute $\boldsymbol{z_{k+1}}$. The nodes compute $\boldsymbol{\nabla{f_n}}(\boldsymbol{z_{k+1}})$ and so on until convergence.  

Note that $\boldsymbol{v_k}$ represents a noisy distorted version of the global momentum-based gradient of $F(\boldsymbol{z_k})$. These effects on the convergence rate will be analyzed in Section \ref{sec:performance}. Also, $\beta$ is a constant stepsize that will be designed in Section \ref{sec:performance}. The pseudocode of the AGMA algorithm is given in Algorithm \ref{alg:AGMA}.

 \begin{algorithm}
  \caption{AGMA algorithm}\label{alg:AGMA}
    \begin{algorithmic}[1]
    \State \textbf{initializing:} $\boldsymbol{\theta_{0}}$, $\boldsymbol{z_{0}}=\boldsymbol{\theta_{0}}$.
    \Repeat{\;(iteration $k=0, 1, ...$):}
    \State \textbf{if} $k\geq 1$ \textbf{then}
    \State \hspace{0.5cm} Each node and the PS update $\boldsymbol{z_{k}}$ according to (\ref{eq:z_k})
    \State \textbf{end}
    \State Each node transmits 
    $y_{n}(\boldsymbol{z_{k}},t)$ according to (\ref{eq:y_n_z_t})
      \State PS receives signal $r_k(t)$ according to (\ref{eq:r_k_t})
     \State PS generates $\boldsymbol{v_k}$ according to (\ref{eq:v_k_definition})
      \State PS updates its estimate $\boldsymbol{\theta_{k+1}}$ according to (\ref{eq:theta_k_1})
      \State PS stores $\boldsymbol{\theta_{k+1}}$ for the next two iterations
      \State PS broadcasts $\boldsymbol{\theta_{k+1}}$ to the nodes
      \State Each node stores $\boldsymbol{\theta_{k+1}}$ for the next two iterations      
    \Until convergence
    \end{algorithmic}
  \end{algorithm}

\section{Performance Analysis}
\label{sec:performance}

In this section, we analyse the performance of the AGMA algorithm. The index $n$ is used for the node index, and $k$ is used for the iterate update at time slot $t_k$. The error (or the excess risk) of GD-type algorithms is commonly defined as the loss in the objective value at iteration $k$ with respect to the optimal value \cite{Nesterov2004IntroductoryLO}:
\begin{equation}
\label{eq:excess_risk}
\displaystyle\mathbb{E}[F(\boldsymbol{\theta_k})]-F(\boldsymbol{\theta^*}),
\end{equation}
where the expectation is taken with respect to the randomness of the generated estimate $\boldsymbol{\theta_k}$ (i.e., the random channel fading and the additive noise in this paper). We are interested in characterizing the rate at which the error decreases with the number of iterations $k$. As commonly adopted in online learning algorithms, linear and sublinear convergence rates are referred to error decay as $O(c^k)$, for $0<c<1$, and $O(1/k^{\epsilon})$, for any $\epsilon > 0$, respectively (i.e., linear and sublinear convergence rates, respectively, on a semi-log plot). 

\subsection{Analyzing AGMA under Strongly Convex Objective with Lipschitz Gradient}
\label{ssec:strongly_lip}

We start by analyzing AGMA under the assumption that $F(\boldsymbol{\theta})$ is strongly convex with strong convexity constant $\mu$, and has Lipschitz gradient with Lipschitz constant $L$, where\footnote{Note that $\mu\leq L$ always holds for strongly convex function with Lipschitz gradient (see Appendix \ref{app:lemmas}). Here, we require slightly stronger assumption that strict inequality would hold.} $\mu<L$. It is known that using momentum in accelerating centralized GD algorithm achieves linear convergence rate under strongly convex with Lipschitz gradient functions\cite{Nesterov}. In the main theorem below we establish a finite-sample error bound of AGMA over noisy fading channels. We show that by designing the algorithm parameters judiciously, AGMA approaches the same best-known linear rate of the centralized accelerated GD as $N$ increases. The momentum stepsize $\eta_k$ is determined by the sequence $\alpha_k$, defined in the theorem below.
\vspace{0.2cm}

\begin{theorem}
\label{th:error_bound_fading_channel}
Consider the system model specified in Section \ref{sec:system}, and strongly convex objective function with Lipschitz gradient as specified in this subsection. Let $\boldsymbol{z_{0}}=\boldsymbol{\theta_{0}}$ for some choice of $\boldsymbol{\theta_{0}}\in\mathbb{R}^{d}$. Choose $\alpha_{0}\in(\sqrt{\frac{\mu}{L}},1)$, and let $\alpha_{k+1}\in(0,1)$ such that it solves the following equation\footnote{It is shown in the analysis in the Appendix that the solution to (\ref{eq:th1_a_2}) is unique on $(0,1)$.}:
\begin{equation}
\label{eq:th1_a_2}
    \alpha_{k+1}^{2} = (1-\alpha_{k+1})\alpha_{k}^{2}+\frac{\mu}{\widetilde{L_\beta}}\alpha_{k+1}. 
\end{equation}
Set $\eta_{k}$ in (\ref{eq:z_k}) to be:  
\begin{equation}
\label{eq:th1_eta_k}
    \eta_{k} = \frac{\alpha_{k}(1-\alpha_{k})}{\alpha_{k+1}+\alpha_{k}^{2}}.
\end{equation}
Let the constant stepsize in \eqref{eq:theta_k_1} satisfy:
\begin{equation}
    \label{eq:step_size}
        0<\beta< \frac{2}{\mu_{h}L}.
\end{equation}
Let $\boldsymbol{\theta^*}$ denote the solution of the optimization problem in (\ref{eq:theta_star}).
Then, for all $k$, the error under the AGMA algorithm is bounded by:
\begin{equation}
\label{eq:th_error_f_equal_strong_convex}
\begin{array}{l}
\displaystyle\mathbb{E}[F(\boldsymbol{\theta_k})]-F(\boldsymbol{\theta^*}) \vspace{0.2cm}\\\hspace{0.3cm}
\displaystyle\leq\left(1-\sqrt{\frac{\mu}{\widetilde{L_\beta}}}\right)^{k}(F(\boldsymbol{\theta_{0}})-F(\boldsymbol{\theta^{*}})+\frac{\gamma_{0}}{2}\|\boldsymbol{\theta_{0}}-\boldsymbol{\theta^{*}}\|^{2})
\vspace{0.2cm}\\\hspace{2cm}
\displaystyle+\sqrt{\frac{\widetilde{L_\beta}}{\mu}}\frac{\beta}{\mu_{h}}\left(\frac{\sigma_{h}^{2}G}{N}+\frac{d\sigma_{w}^{2}}{E_{N}N^{2}}\right),
\end{array}
\end{equation}
where 
\begin{eqnarray}
\label{eq:th1_gamma_0}
\gamma_{0}\triangleq\frac{\alpha_{0}(\alpha_{0}L-\mu)}{1-\alpha_{0}},   
\end{eqnarray}
and
\begin{equation}
    \widetilde{L_\beta} \triangleq \frac{1}{\beta\Big(\frac{2}{\mu_{h}}-\beta L\Big)\mu_h^2}=\frac{1}{\tilde{\beta}\Big(2-\tilde{\beta} L\Big)},\label{th:L_beta}
\end{equation}
\end{theorem}
where $0<\tilde{\beta}\triangleq\beta\mu_h<2/L$.

The proof is given in Appendix \ref{app:proof1}. 

Throughout the proof, we use recursion auxiliary functions to bound the error. These functions take into account both the gradient descent update step and the momentum step. In order to achieve the desired bound, we apply the control sequence of the momentum, $\alpha_k$, that can be interpreted as a memory factor. Specifically, the sequence is of the from: $\phi_{k+1}(\theta)=(1-\alpha_k)\phi_k(\theta)+\alpha_k A_{z_k}(\theta)$, where $A_{z_k}(\theta)$ denotes the second order approximation of $F(\boldsymbol{\theta})$ around the momentum update $\boldsymbol{z_{k}}$, depending on the noisy distorted gradient, $\boldsymbol{v_{k}}$, and the guarding term $\epsilon_N$ used to bound the noise effect.
A discussion of the results implied by Theorem \ref{th:error_bound_fading_channel} is given next.\vspace{0.2cm}
\subsubsection{Effect of the divergence with respect to the initial estimate model on the error} The first term that affects the error is the following divergence with respect to the initial estimate model: $F(\boldsymbol{\theta_{0}})-F(\boldsymbol{\theta^{*}})+\frac{\gamma_{0}}{2}\|\boldsymbol{\theta_{0}}-\boldsymbol{\theta^{*}}\|^{2}$. It is known that under strongly convex objective function with Lipschitz gradient, a centralized accelerated GD (i.e., without gradient distortion due to fading effect: $\mu_h=1, \sigma_h^2=0$, and noise-free channel: $\sigma_w^2=0$) achieves the best linear convergence rate when using stepsize $\beta=\frac{1}{L}$ \cite{Nesterov2004IntroductoryLO}: 
\begin{equation}
\begin{array}{l}
\mathbb{E}[F(\boldsymbol{\theta_k})]-F(\boldsymbol{\theta^*}) \leq \vspace{0.2cm} \\\hspace{0.3cm}
\left(1-\sqrt{\frac{\mu}{L}}\right)^{k}(F(\boldsymbol{\theta_{0}})-F(\boldsymbol{\theta^{*}})+\frac{\gamma_{0}}{2}\|\boldsymbol{\theta_{0}}-\boldsymbol{\theta^{*}}\|^{2}).
\end{array}
\end{equation}
It is interesting to notice that setting $\beta=\frac{1}{\mu_h L}$ in AGMA yields $\widetilde{L_\beta}=L$ which results in the same linear convergence rate due to the divergence with respect to the initial estimate model as in the centralized accelerated GD algorithm.

\subsubsection{Effect of the channel fading and additive noise on the error}

The second term that affects the error in the strongly convex case, given by: $T_{2,strongly-convex}\triangleq\sqrt{\frac{\widetilde{L_\beta}}{\mu}}\frac{\beta}{\mu_{h}}\frac{\sigma_{h}^{2}G}{N}$, is due to the momentum-based gradient distortion caused by amplifying each local momentum-based gradient by a different random channel gain (when $\sigma_h^2>0$). Noting that we can write $\beta=\tilde{\beta}/\mu_h$, where $0<\tilde{\beta}<2/L$ yields: $T_{2,strongly-convex}=\sqrt{\frac{\widetilde{L_\beta}}{\mu}}\frac{\tilde{\beta}G}{N}\left(\frac{\sigma_{h}}{\mu_{h}}\right)^2=\sqrt{\frac{\widetilde{L_\beta}}{\mu}}\frac{\tilde{\beta}G}{N}\cdot\left(\mbox{CV}_h\right)^2$, where $\mbox{CV}_h\triangleq \frac{\sigma_{h}}{\mu_{h}}$. Note that the channel fading effect on this term is reflected by the standard deviation-to-mean ratio of the channel gain, known as the channel coefficient of variation (CV), which measures the dispersion of the channel gain distribution. As the users experience wireless channels with a larger dispersion, i.e., larger $\mbox{CV}_h$, the distortion of the global gradient increases, which is expected to decrease the performance, as supported by the theoretical analysis. Increasing the number of users $N$, diminishes the distortion effect in the error bound with rate $1/N$ (as can be seen in the analysis). Thus, AGMA approaches the linear convergence rate of the centralized accelerated GD algorithm.

The third term of the error bound, given by: $T_{3,strongly-convex}\triangleq\sqrt{\frac{\widetilde{L_\beta}}{\mu}}\frac{\beta}{\mu_{h}}\frac{d\sigma_{w}^{2}}{E_{N}N^{2}}$, is due to the additive channel noise at the receiver, and the channel gain mean. This term can be written as: $T_{3,strongly-convex}=\sqrt{\frac{\widetilde{L_{\beta}}}{\mu}}\frac{\tilde{\beta} d}{\mu_{h}^2N^{2}}\frac{\sigma_{w}^{2}}{E_{N}}=\sqrt{\frac{\widetilde{L_{\beta}}}{\mu}}\frac{\tilde{\beta} d}{\mu_{h}^2N^{2}}\frac{1}{\mbox{SNR}_N}$, where $\mbox{SNR}_N\triangleq \frac{E_{N}}{\sigma_{w}^{2}}$. Note that increasing the SNR, decreases the error effect of the additive noise, as expected. Similarly, the channel fading effect on this term is reflected by the channel gain mean square. Having a larger channel gain mean decreases the error. It can be seen that this effect decreases as the term $E_{N}N^{2}$ increases. As a result, we can set the transmission power to $E_N=\Omega\left(N^{\epsilon-2}\right)$ to eliminate this term as $N$ increases. Note that the scaling of the transmission energy is properly defined for aggregation, depending on the number of nodes $N$. Improving the learning accuracy and at the same time decreasing the transmission energy has a price, as it requires to increase the number of nodes $N$ that participate in the learning task.

It should be noted that there are many systems that allow controlling the transmission power used to optimize the network performance, as long as power constraints determined by the physical system or regulation requirements are met. Common examples are cognitive radio and mesh networks, where the transmission power is often adjusted to reduce the interference level. Also, in sensor networks, transmission power control plays a key role in the network operation to maximize the network lifetime for example.

\subsection{Analyzing AGMA under Convex Objective with Lipschitz Gradient}
\label{ssec:convex_lip}

In this section we relax the strong convexity assumption, and require that $F(\boldsymbol{\theta})$ would be convex only (but still has Lipschitz gradient). It is well known that linear convergence cannot be achieved in this case even in the centralized setting, but only sub-linear convergence rate. Furthermore, it is known that using momentum in accelerating centralized GD algorithm improves the convergence rate from $1/k$ (under standard GD iterates) to $1/k^{2}$ with acceleration\cite{Nesterov}. Interestingly, we show that for all $k=1, 2, ..., \lfloor N^{1-\epsilon}\rfloor$ for some $\epsilon>0$, using AGMA's acceleration approaches $1/k^{2}$ convergence rate over noisy fading channels as $N$ increases.
\vspace{0.2cm}

\begin{theorem}
\label{th:error_bound_fading_channel_convex}
Consider the system model specified in Section \ref{sec:system}, and convex objective function with Lipschitz gradient as specified in this subsection. Let $\boldsymbol{z_{0}}=\boldsymbol{\theta_{0}}$ for some choice of $\boldsymbol{\theta_{0}}\in\mathbb{R}^{d}$. Choose $\alpha_{0}\in(0,1)$, and let $\alpha_{k+1}\in(0,1)$ such that it solves the following equation\footnote{It is shown in the analysis in the Appendix that the solution to (\ref{eq:th2_a_2}) is unique on $(0,1)$.}:
\begin{equation}
\label{eq:th2_a_2}
    \alpha_{k+1}^{2} = (1-\alpha_{k+1})\alpha_{k}^{2}.
\end{equation}
Set $\eta_{k}$ in (\ref{eq:z_k}) as in \eqref{eq:th1_eta_k}, and let the constant stepsize in \eqref{eq:theta_k_1} as in \eqref{eq:step_size}. Let $\boldsymbol{\theta^*}$ denote the solution of the optimization problem in (\ref{eq:theta_star}). Let $k_0=\lfloor N^{1-\epsilon}\rfloor$. Then, for all $k=1, 2, ..., k_0$, the error under the AGMA algorithm is bounded by:
\begin{equation}
\label{eq:th_error_f_equal_convex1}
\begin{array}{l}
\displaystyle 
\mathbb{E}[F(\boldsymbol{\theta_k})]-F(\boldsymbol{\theta^*}) \vspace{0.2cm}\\\hspace{0.3cm}
\displaystyle \leq  \frac{4\widetilde{L_\beta}}{(2\sqrt{\widetilde{L_\beta}}+k\sqrt{\gamma_{0}})^{2}}(F(\boldsymbol{\theta_{0}})-F(\boldsymbol{\theta^{*}})+\frac{\gamma_{0}}{2}\|\boldsymbol{\theta_{0}}-\boldsymbol{\theta^{*}}\|^{2})\vspace{0.2cm}\\\hspace{2cm} \displaystyle +\frac{\beta}{\mu_{h}}\left(\frac{\sigma_{h}^{2}G}{N^\epsilon}+\frac{d\sigma_{w}^{2}}{E_{N}N^{1+\epsilon}}\right),
\end{array}
\end{equation}
where $\gamma_{0}\triangleq\frac{\alpha_{0}^{2}L}{1-\alpha_{0}}$ (since $\mu=0$ in \eqref{eq:th1_gamma_0}), and $\widetilde{L_\beta}$ is given in \eqref{th:L_beta}.
\end{theorem}

The proof is given in Appendix \ref{app:proof2}. 
Note that relaxing the strongly-convex assumption, and assuming only convex objective function is known to decrease the performance in terms of convergence order. A discussion of the results implied by Theorem \ref{th:error_bound_fading_channel_convex} is given next. 

\subsubsection{Implementation of AGMA for all $k$} Note that the bound in Theorem \ref{app:proof2} is valid only for $k=1, 2, .., \lfloor N^{1-\epsilon}\rfloor$. The effect of diverged error in accelerated GD algorithms is well known when handling noisy gradients, and heuristic restarted methods that avoid momentum are often adopted when $k$ is large (see e.g., \cite{wang2020scheduled},\cite{Adaptive_Restart} and references therein). To the best of our knowledge, our theoretical analysis provides the first result that guarantees convergence analytically in accelerated GD over noisy fading MAC. Specifically, for all $k=1, 2, .., \lfloor N^{1-\epsilon}\rfloor$ using AGMA improves the convergence rate from $1/k$ to $1/k^2$, and for all $k>\lfloor N^{1-\epsilon}\rfloor$ one can set the momentum constant to $\eta_k=0$ to guarantee $1/k$ convergence rate by standard GD with noisy distorted gradients as was shown in \cite{sery2020analog}. In the simulation results, we set $k_0$ such that the error bound is minimized which demonstrated better performance by AGMA as compared to existing methods.

\subsubsection{Effect of the divergence with respect to the initial estimate model on the error} 
Similar to the strongly-convex case, in the convex case considered here as well the first term that affects the error is the following divergence with respect to the initial estimate model: $F(\boldsymbol{\theta_{0}})-F(\boldsymbol{\theta^{*}})+\frac{\gamma_{0}}{2}\|\boldsymbol{\theta_{0}}-\boldsymbol{\theta^{*}}\|^{2}$. It is known that under convex objective function with Lipschitz gradient, a centralized accelerated GD (i.e., without gradient distortion due to fading effect: $\mu_h=1, \sigma_h^2=0$, and noise-free channel: $\sigma_w^2=0$) achieves an improved sub-linear convergence rate when using stepsize $\beta=\frac{1}{L}$ \cite{Nesterov2004IntroductoryLO}: 
\begin{equation}
\begin{array}{l}
\mathbb{E}[F(\boldsymbol{\theta_k})]-F(\boldsymbol{\theta^*}) \vspace{0.2cm}\\\hspace{0.3cm}
\displaystyle \leq  \frac{4L}{(2\sqrt{L}+k\sqrt{\gamma_{0}})^{2}}(F(\boldsymbol{\theta_{0}})-F(\boldsymbol{\theta^{*}})+\frac{\gamma_{0}}{2}\|\boldsymbol{\theta_{0}}-\boldsymbol{\theta^{*}}\|^{2}).
\end{array}
\end{equation}
It is interesting to notice that setting $\beta=\frac{1}{\mu_h L}$ in AGMA in the convex case as well yields $\widetilde{L_\beta}=L$ which results in the same sub-linear convergence rate due to the divergence with respect to the initial estimate model as in the centralized accelerated GD algorithm.

\subsubsection{Effect of the channel fading and additive noise on the error}
The second term that affects the error in the convex case, given by: $T_{2,convex}\triangleq\sqrt{\frac{\widetilde{L_\beta}}{\mu}}\frac{\tilde{\beta}G}{N^{\epsilon}}\cdot\left(\mbox{CV}_h\right)^2$, is due to the momentum-based gradient distortion caused by amplifying each local momentum-based gradient by a different random channel gain (when $\sigma_h^2>0$). Nevertheless, as the number of nodes $N$ increases the distortion effect decreases and AGMA approaches the sub-linear convergence rate of the centralized accelerated GD for $k=1, 2, ..., \lfloor N^{1-\epsilon}\rfloor$. In this case, however, the distortion effect in the error bound diminishes only with rate $1/N^{\epsilon}$ (as can be seen in the analysis).

The third term that affects the error bound: $T_{3,convex}\triangleq\frac{\tilde{\beta} d}{\mu_{h}^2N^{1+\epsilon}}\frac{\sigma_{w}^{2}}{E_{N}}=\frac{\tilde{\beta} d}{\mu_{h}^2N^{1+\epsilon}}\frac{1}{\mbox{SNR}_N}$ is due to the additional channel noise at the receiver (when $\sigma_w^2>0$). As expected, larger SNR, or larger channel gain mean decreases the error effect of the additive noise. Also, it can be seen that in this case this term decreases as the term $E_{N}N^{1+\epsilon}$ increases (i.e., slower rate than in the strongly-convex case). As a result, we can set the transmission power to $E_N=\Omega\left(N^{-1-\epsilon'}\right)$, for some $0<\epsilon'<\epsilon$, to eliminate this term as $N$ increases.

\section{Simulation Results}
\label{sec:sim}

We now provide numerical examples to illustrate the performance of AGMA in two different settings. In the first setting, we simulated a federated learning task used to predict a release year of a song from its audio features. We used real-data, the popular Million Song Dataset \cite{Bertin-Mahieux2011}, and distributed it among a large number of edge devices with the goal of training the global predictor. In the second setting, we simulated a distributed learning task for estimation in radar systems. The model consists of array of radars in order to capture evidence of free electrons in the ionosphere. We used the popular Ionosphere real dataset collected by a radar system in Goose Bay, Labrador \cite{Dua:2019}.

We compared AGMA with the following gradient-based learning algorithms that  demonstrated very good performance in federated learning tasks recently in the literature: (i) The error compensated entry-wise scheduled analog  distributed stochastic gradient descent (ECESA-DSGD) algorithm \cite{amiri2020federated}, in which each node transmits the gradient at each iteration only if the channel state exceeds a certain threshold. The signal is normalized by the channel state. (ii) The Gradient-Based Multiple Access (GBMA) algorithm \cite{sery2020analog}, in which all nodes transmit the local gradient to the PS over the fading noisy MAC without power control. (iii) The FDM-GD algorithm, in which each node is allocated a dedicated orthogonal channel for transmission. The PS receives all signals and calculates the mean signal. This scheme was widely used in federated learning applications (see e.g., \cite{konevcny2016federated}). (iv) The FDM-accelerated gradient descent (FDM-AGD) algorithm, which is similar to FDM-GD, but uses momentum-based gradient in iterate updates to accelerate convergence. We set the transmission parameters of the algorithms such that the average transmitted power per node is equal for all algorithms.  

\subsection{Federated Learning for Predicting a Release Year of a Song}
\label{ssec:real}

We start by examining the federating learning task of predicting a release year of a song from its audio attributes. We used the Million Song Dataset \cite{Bertin-Mahieux2011}, which contains songs which are mostly western, commercial tracks ranging from 1922 to 2011. Each song is tagged with the release year (i.e., the label) and $90$ audio attributes (i.e., the input vector).

We used linear least squares regression loss $f_n(\boldsymbol{\theta}) = \frac{1}{2|\mathcal{D}_n|} \sum_{d\in \mathcal{D}_n}{\left(\boldsymbol{x}_{n,d}^T\boldsymbol{\theta}-y_{n,d}\right)^2}$, where $\mathcal{D}_n$ is a local data set of node $n$, which satisfies the convexity and Lipschitz gradient requirements in Subsection \ref{ssec:convex_lip}. 

\begin{figure}[htbp] 
\begin{center}
    \subfigure[The error as a function of the number of iterations for linear least squares loss.]{\scalebox{0.4}
    {
      \label{fig:1a}
      \includegraphics[height=0.4\textheight,width=0.9\textwidth]{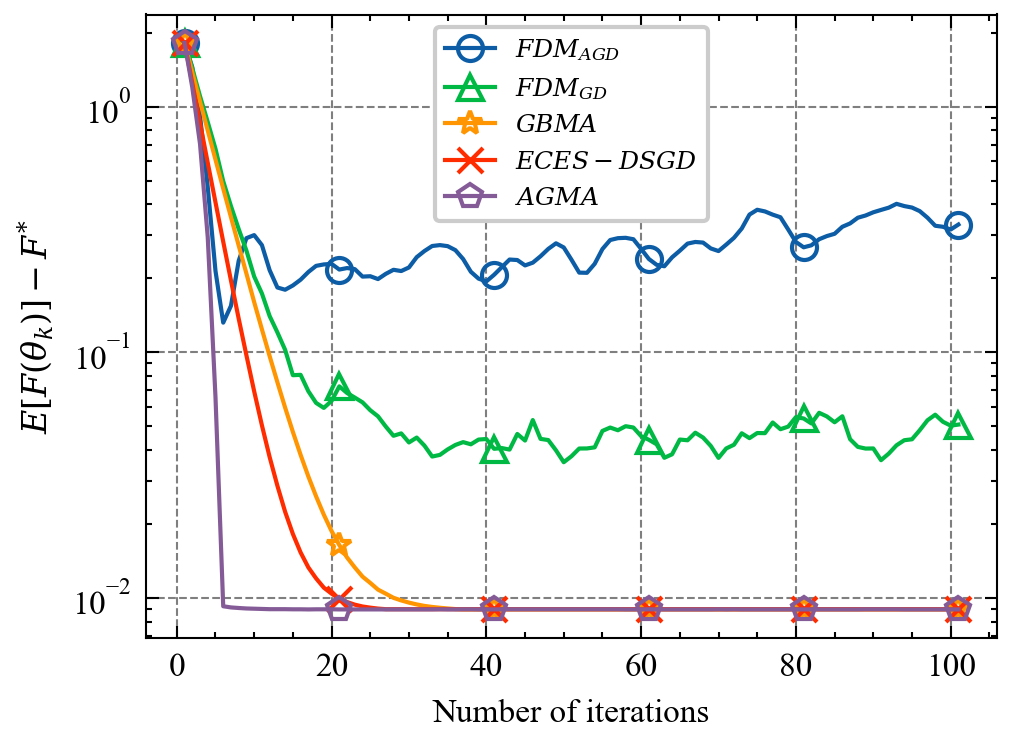}
    }}
    \subfigure[The error after 10 iterations as a function of the number of nodes.  ]{\scalebox{0.4}
    {
      \label{fig:1c}
      \includegraphics[height=0.4\textheight,width=0.9\textwidth]{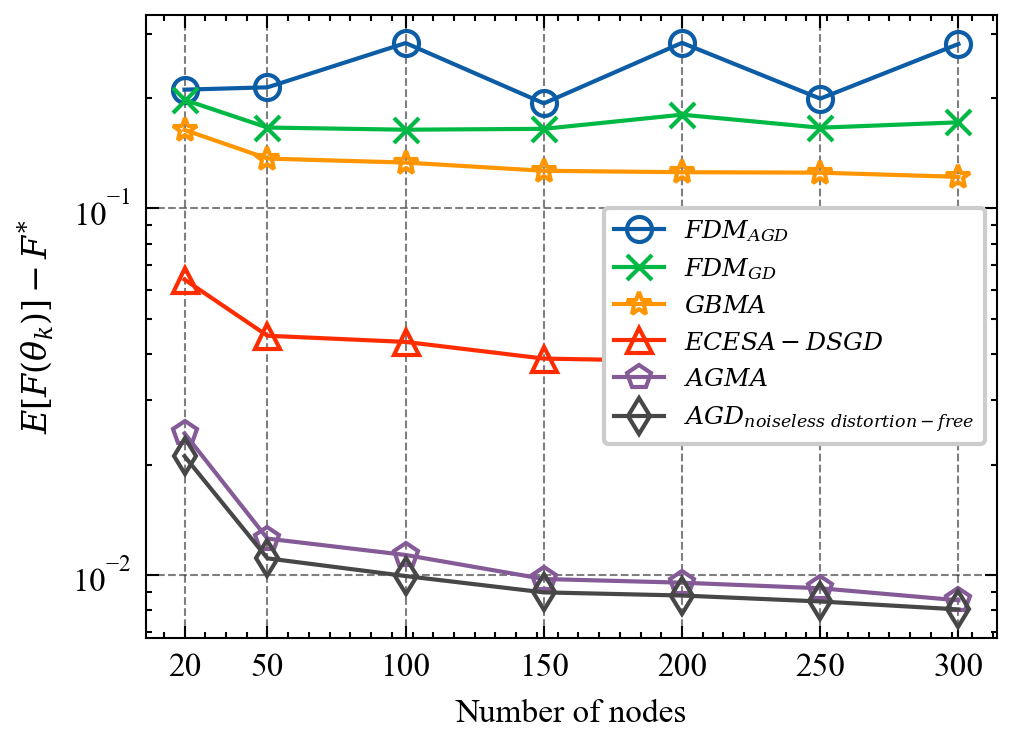}
    }}
        \subfigure[The error as a function of the number of iterations for linear least squares log-loss.]{\scalebox{0.4}
    {
      \label{fig:non-convex}
      \includegraphics[height=0.4\textheight,width=0.9\textwidth]{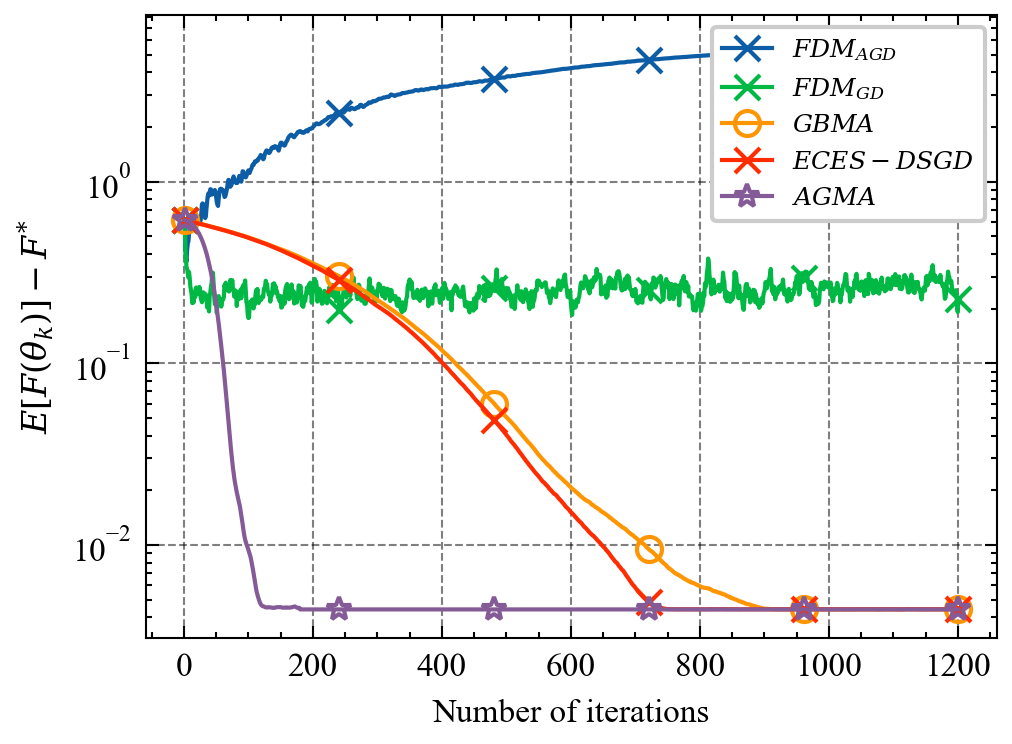}
    }}
    \caption{Algorithm comparison for the federated learning setting of predicting a release year of a song. In the first figure, the error is presented as a function 
    of the number of iterations for a linear least squares loss function (i.e., a convex loss function) with $N=150$, $\sigma_h^2=0.5$, $\sigma_w^2=1, E_N=1$. 
    The second figure demonstrates the observed error after 10 iterations, where AGD over noiseless distortion-free channel serves as a benchmark on the performance. We set: $\sigma_h^2 = 0.8, E_N=1$, $\sigma_w^2=1$. In the third figure, the error is presented as a function of the number of iterations for a linear least squares log-loss  function (i.e., a non-convex loss function) with $N=150$, $\sigma_h^2=0.5$, $\sigma_w^2=1, E_N=1$.}
  \label{fig:1}
\end{center}
  \end{figure}

In Fig.\ref{fig:1a} we present the performance comparison of the achievable average error. Recall that each iteration presents a single communication round, where FDM-GD and FDM-AGD require $N$ times higher bandwidth requirement as compared to GBMA, ECESA-DSGD, and AGMA that transmit over MAC in each round. Thus, improving the convergence rate translates to reducing the bandwidth and latency requirements. It can be seen that AGMA significantly outperforms all other algorithms and achieves the smallest error in just $5$ iterations. By contrast, ECESA-DSGD and GBMA reach the smallest error in more than $20$ iterations. FDM-GD and FDM-AGD perform worse as well. In Fig. \ref{fig:1c} we present the performance of the algorithms after 10 iterations. The AGD over noiseless distortion-free channel serves as a benchmark for performance to present the optimality gap of the algorithms. It can be seen that AGMA performs the best and almost approaches the AGD over noiseless distortion-free channel. In Fig. \ref{fig:non-convex} we examine the performance of the algorithms when the loss function is not convex, and consequently the theoretical conditions are not met. We used linear least squares log-loss\cite{ramamohanaraoalgorithms}:  $f_{n}(\boldsymbol{\theta}) = \frac{1}{2|\mathcal{D}_n|} \sum_{d\in \mathcal{D}_n}{\log{\left((\boldsymbol{x}_{n,d}^T\boldsymbol{\theta}-y_{n,d})^2+1\right)}}$. 
It can be seen that AGMA achieves the best convergence rate in this case as well, which further demonstrates the advantage of using the proposed AGMA in practice, even when the theoretical conditions are not met.

\begin{figure}[htbp] 
\begin{center}
    \subfigure[The error of AGMA for different values of $N$.]{\scalebox{0.4}
    {
      \label{fig:3a}
      \includegraphics[height=0.4\textheight,width=0.9\textwidth]{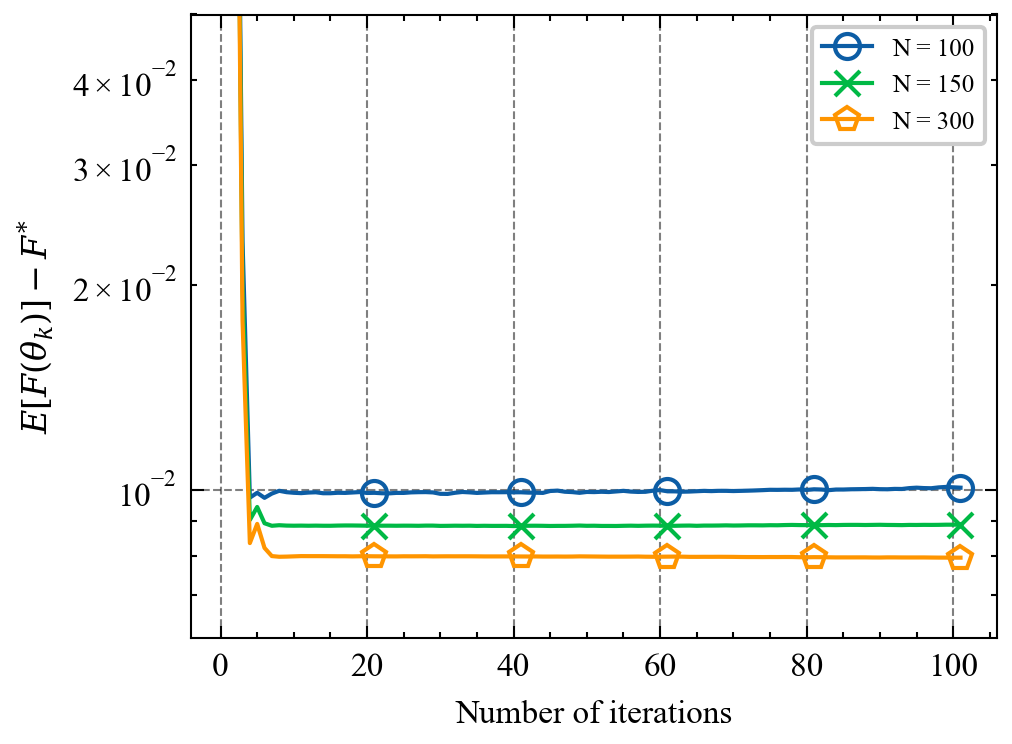}
    }}
    \subfigure[The error of AGMA as a function of the power coefficient $E_N$ after 10 iterations.]{\scalebox{0.4}
    {
      \label{fig:3b}
      \includegraphics[height=0.4\textheight,width=0.9\textwidth]{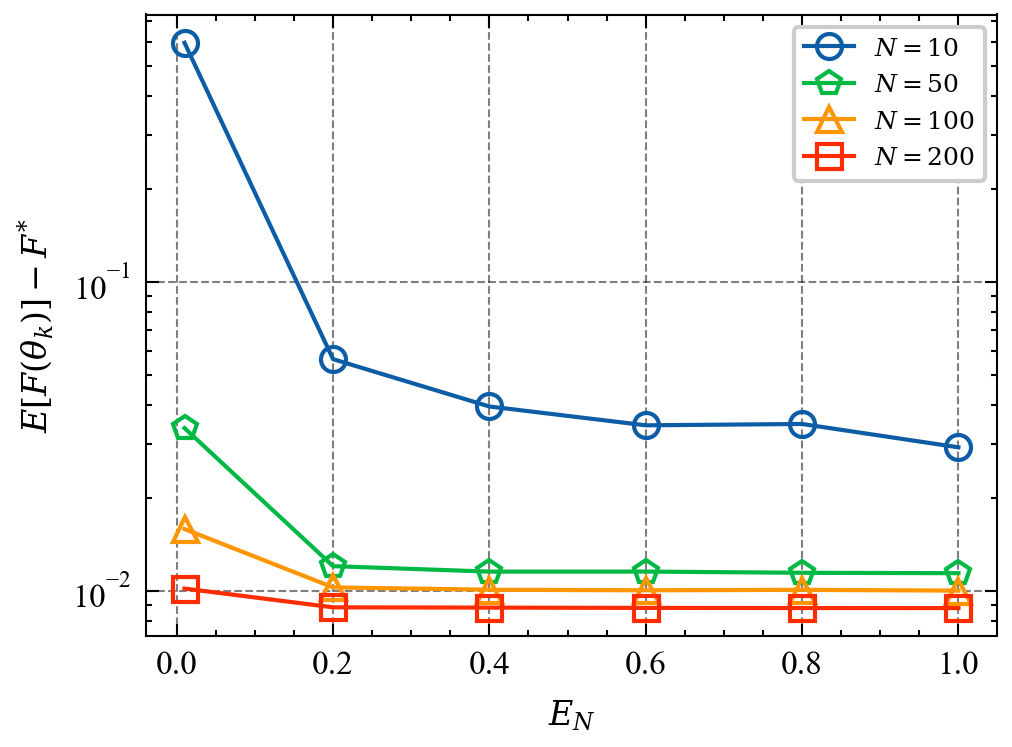}
    }} 
    \subfigure[The error of AGMA for different values of stepsize $\beta_f$.]{\scalebox{0.4}
    {
      \label{fig:3c}
      \includegraphics[height=0.4\textheight,width=0.9\textwidth]{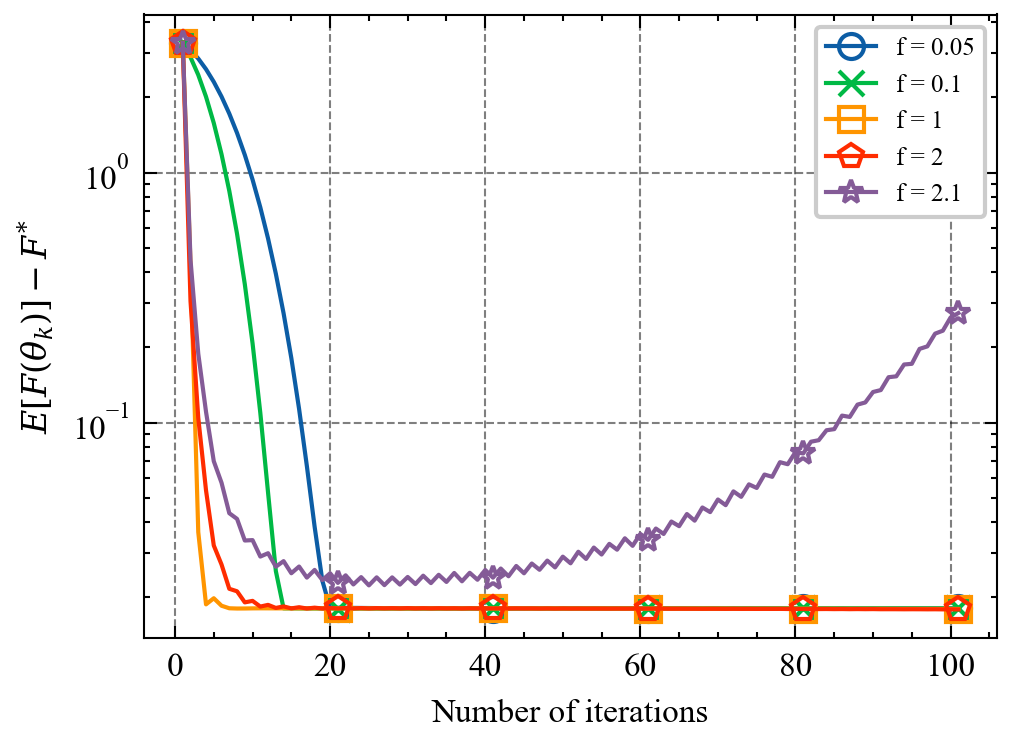}
    }}
    \subfigure[The error of AGMA for different values of $\alpha_0$ that controls the momentum.]{\scalebox{0.4}
    {
      \label{fig:3d}
      \includegraphics[height=0.42\textheight,width=0.9\textwidth]{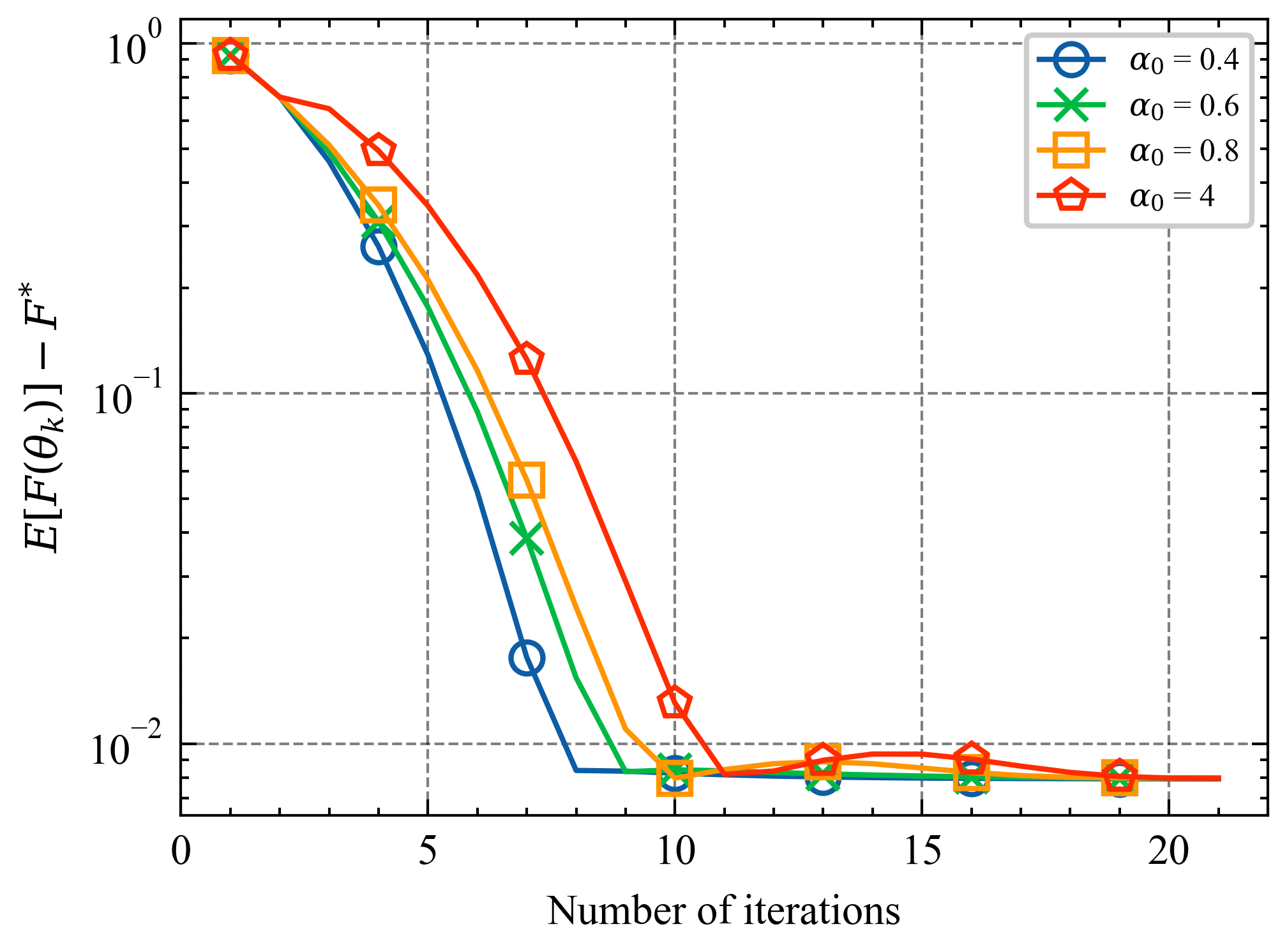}
    }} \\
    \caption{The performance of AGMA under various simulation parameters, with $\sigma_h^2=0.5, \sigma_w^2=1$. In Fig. \ref{fig:3a}, the performance is presented for $N=100, 150, 300, E_N=1$. In Fig. \ref{fig:3b}, the error of AGMA is presented after $10$ iterations as a function of the power coefficient $E_N$ (fixed for all $N$). In Fig. \ref{fig:3c}, the error of AGMA is presented for different values of stepsize parameter $\beta_f=f/\mu_h L$, with $E_N=1, N=150$. In Fig.  \ref{fig:3d}, the error is presented for various momentum parameters, determined by $\alpha_0$, with $E_N=1, N=150$.}
\end{center}
  \end{figure}

In Fig. \ref{fig:3a} we simulated AGMA for $N=100, 150, 300$. It can be seen that the error decreases as $N$ increases. This is because the additional terms due to the fading channel and additive noise are mitigated as $N$ increases as observed by the theoretical study. 

In Fig. \ref{fig:3b}, the error of AGMA is presented after $10$ iterations as a function of the power coefficient $E_N$ (fixed for all $N$).
 It can be seen that the error decreases with $E_N$, as expected. Nevertheless, for large $E_N$, the approximation error which decreases with the iterations dominates the error, and thus the error does not decrease with $E_N$. Similarly, it can be seen that the error decreases with $N$, but the improvement mitigates again for large $N$ for the same reason. In Fig. \ref{fig:3c} we tested AGMA under different stepsize values, given by $\beta_f = \frac{f}{\mu_h L}$. As derived in the theoretical analysis, the stepsize that minimizes the error bound is given by: $\beta = \frac{1}{\mu_h L}$, i.e., $f=1$. Furthermore, note that $f>2$ is outside the theoretical convergence range \eqref{eq:step_size}. Interestingly, we indeed achieve the fastest convergence for $f=1$, where the algorithm diverges for $f=2.1$. In Fig. \ref{fig:3d} we simulated AGMA with various momentum parameters determined by $\alpha_0$. The theoretical analysis guarantees convergence for values of $\alpha_0$ in $\alpha_0\in(0,1)$. Indeed, we observed faster convergence for values inside this range. Nevertheless, it can be seen that selecting $\alpha_0=2$ still converges in this experiment.

\subsection{Detecting Radar Errors using a Wireless Network}

In the second setting we simulated a binary classification model for distributed sensing system. We used the popular Ionosphere real dataset collected by a radar system in Goose Bay, Labrador available by UCI Machine Learning Repository \cite{Dua:2019}. The Ionosphere dataset contains radar data to detect free electrons in the ionosphere. "Good" radar returns are those showing evidence of some type of structure in the ionosphere. "Bad" returns are those that do not; their signals pass through the ionosphere. 

We assume distributed radars that transmit their measurements to the PS to learn a shared model for classifying the radar signals. The network consists of $N=150$ radars, each holds local sample data. We used regularized logistic regression loss for each radar $n$:

    $f_{n}(\boldsymbol{\theta}) = \log{\left(1+\exp{(-y_{n}\boldsymbol{\theta}^T\boldsymbol{x_{n}})}\right)}+\frac{\lambda}{2}\|\boldsymbol{\theta}\|_{2}^{2}$,

which satisfies the strong-convexity and Lipschitz gradient requirements in Subsection \ref{ssec:strongly_lip}. We set $\lambda = 0.1$.

\begin{figure}[H] 
\begin{center}
    \subfigure[The error as a function of the number of iterations.]{\scalebox{0.4}
    {
      \label{sim:fig_radar_a}
      \includegraphics[height=0.4\textheight,width=0.9\textwidth]{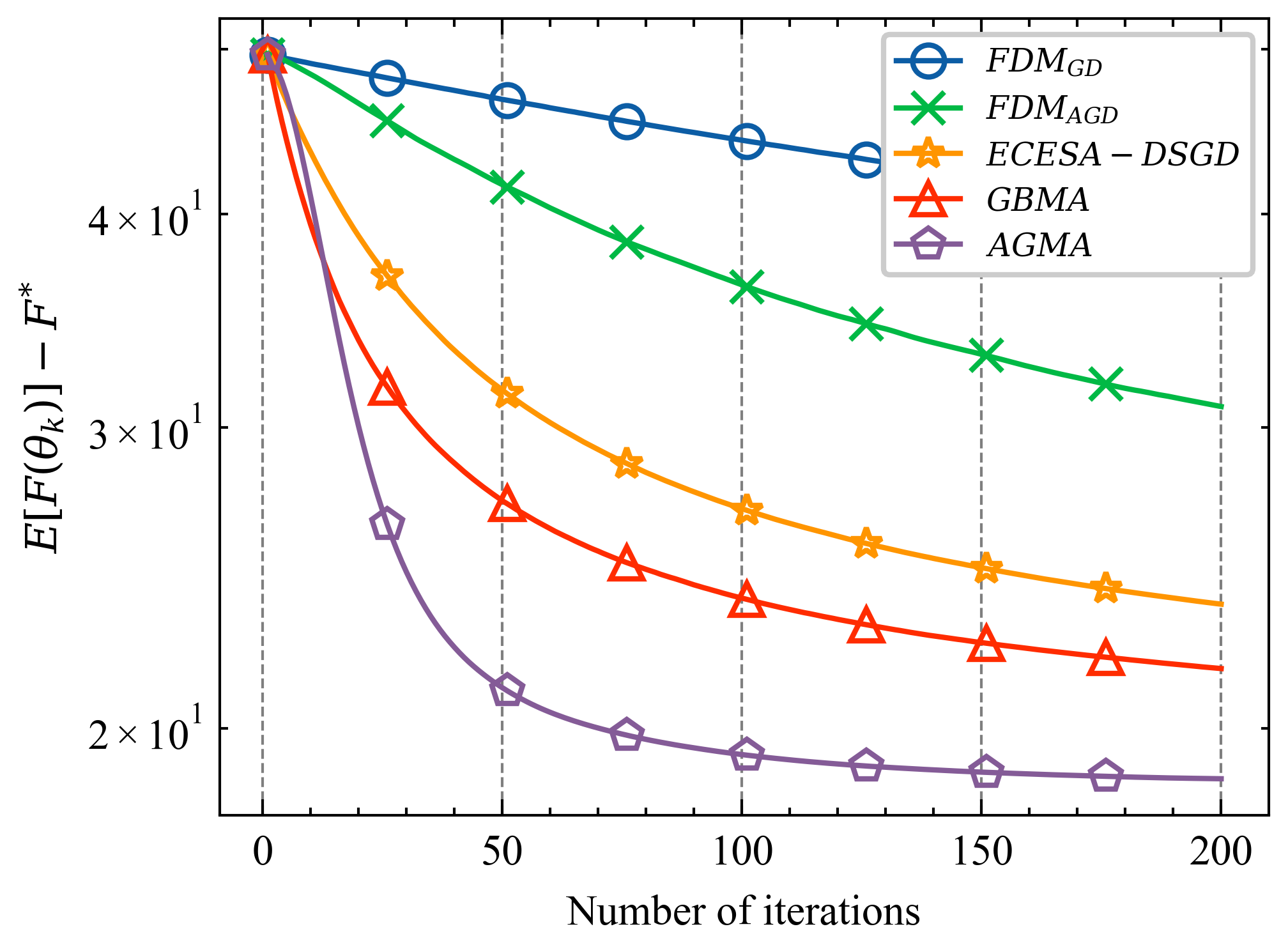}
    }}
    \subfigure[The error of AGMA as a function of the transmission power coefficient $E_N$.]{\scalebox{0.4}
    {
      \label{sim:fig_radar_b}
      \includegraphics[height=0.4\textheight,width=0.9\textwidth]{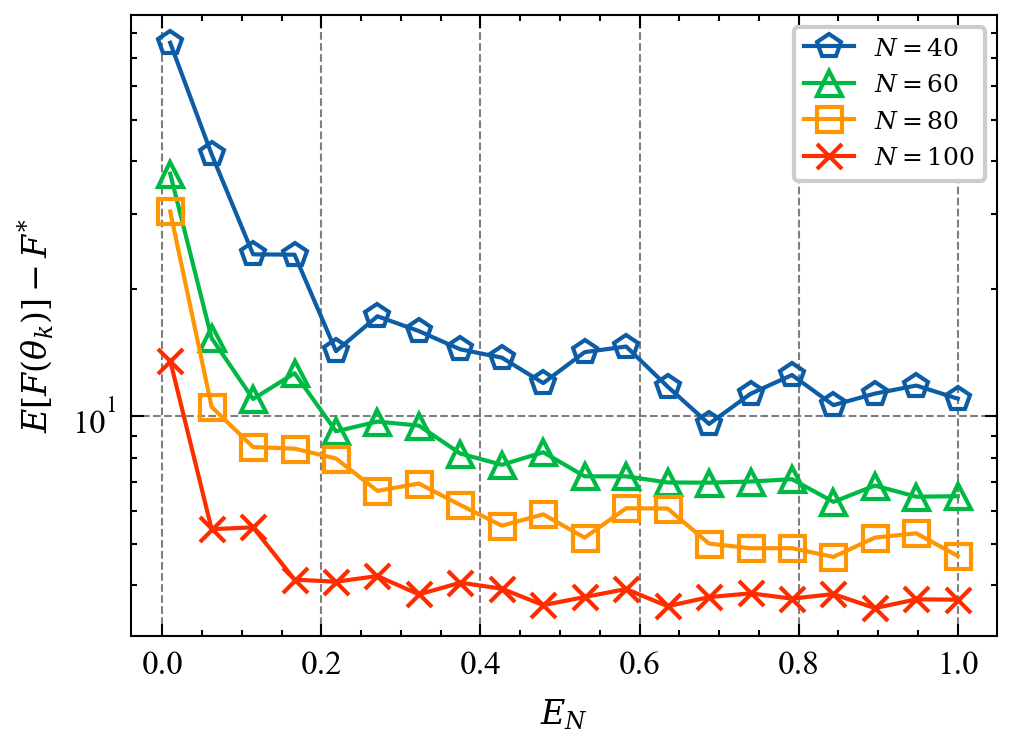}
    }}
    \subfigure[The error of AGMA for different values of channel parameters. ]{\scalebox{0.4}
    {
      \label{sim:fig_radar_c}
      \includegraphics[height=0.4\textheight,width=0.9\textwidth]{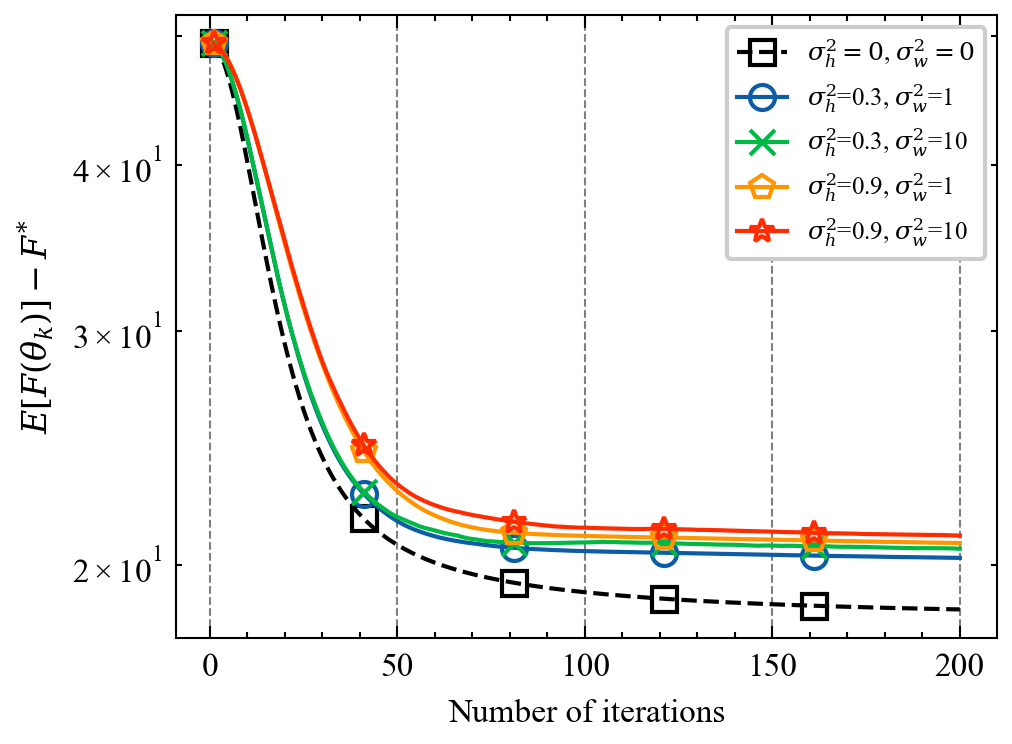}
    }}
    \caption{Algorithm comparison for detecting radar errors using a wireless network. In the first figure, an algorithm comparison is presented as a function of the number of iterations for $N=150$, $\sigma_h^2=0.3$, $\sigma_w^2=0.2$, $E_N=1$. In the second figure, AGMA performance is presented as a function of the transmission power coefficient $E_N$ for different numbers of nodes. In the third figure, AGMA performance is presented as a function of the number of iterations for different values of channel parameters $\sigma_h^2$ and $\sigma_w^2$, with $N=150$, $E_N=1$ .}
  \label{fig:1}
\end{center}
  \end{figure}

In Fig. \ref{sim:fig_radar_a} we present the performance comparison of the achievable average error. It can be seen again that AGMA significantly outperforms all other algorithms. For example, achieving AGMA's error at iteration $50$ requires more than $200$ iterations by GBMA, where the other algorithms perform worse. In Fig. \ref{sim:fig_radar_b} we simulated AGMA for $E_N=N^{\epsilon-2}$, where $\epsilon=1, 2$. Theoretically, as stated in Theorem \ref{th:error_bound_fading_channel}, we can preserve the convergence rate and still keep the total power in the network fixed by setting $E_N=1/N$ as $N\rightarrow\infty$. In the setting here we show that we obtain this property using only $N=150$, which results in significant power savings.
In Fig. \ref{sim:fig_radar_b} we present the error of AGMA as a function of the transmission power coefficient $E_N$. It can be seen that increasing the power improves the performance. Nevertheless, for large values of $E_N$ the approximation error which decreases with the iterations dominates the error, and thus the error does not decrease with $E_N$. In Fig. \ref{sim:fig_radar_c} we simulated AGMA under various channel parameters. It can be seen that increasing the channel or noise variance deteriorates the performance, as expected, and supported by the theoretical analysis. The case of $\sigma_w^2=0$, $\sigma_h^2=0$ serves as a benchmark for comparison to demonstrate the optimality gap as compared to AGMA over a noiseless distortion-free channel. It can be seen that as the number of iterations increases, the error does not decrease in the presence of noise and fading channels, as the estimation error that depends on them dominates the error, and the optimality gap increases. Mitigating this effect requires to increase the number of nodes $N$. These results support the theoretical analysis.

\subsection{Federated Learning for classification of handwritten digits.} 
\label{ssec:realNN}
\begin{figure}[htbp]
\begin{center}{\scalebox{0.75}
  { \includegraphics[height=0.22\textheight,width=0.5\textwidth]{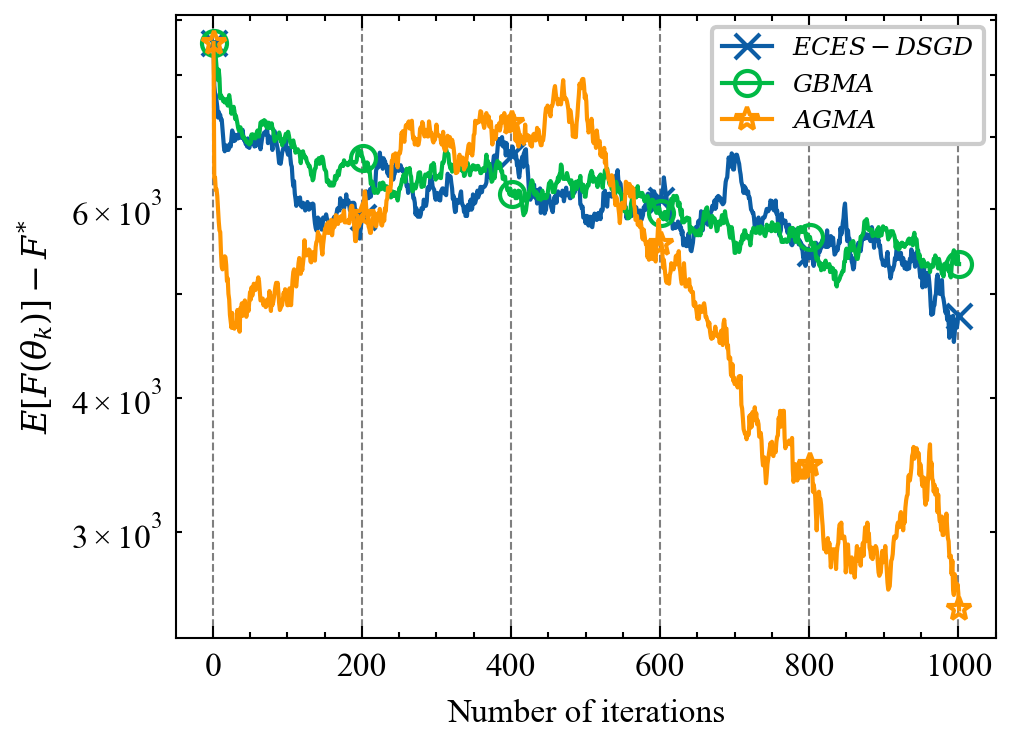}}
  }
   \caption{Simulation results for a federated learning task of classification of handwritten digits via NN. An algorithm comparison of ECES-DSGD, GBMA, and the proposed AGMA is presented with $N=200, \sigma_h^2=0.5, \sigma_w^2=0.1$ and $E_N=1$.}
  \label{sim:fig_NN}
  \end{center}
  \end{figure}

Finally, we consider the MNIST dataset \cite{mnist} for classification of handwritten digits using a neural network (NN). The NN consists of 5 dense layers with ReLU and Leaky ReLU activation, and one drop-out layer. This model has $75,434$ trainable variables, the input is of the form of 28x28 image, and the output has $10$ different categories (digits) classified by softmax activation. This setting does not meet the theoretical convexity conditions. The training is done by a federated learning setting, where the nodes transmit the local gradients to the server, which are aggregated over the fading MAC. We simulated the three algorithms that achieved the best performance in the previous experiments, namely ECES-DSGD, GBMA, and the proposed AGMA. As can be seen in Fig. \ref{sim:fig_NN}, for the first 500 iterations, AGMA does not outperform ECES-DSGD, and GBMA. However, AGMA achieves significant improvement and steep learning curve as the number of iterations increases. These results further demonstrate the significance of using AGMA over noisy fading MAC, even when the theoretical conditions are not met.

\section{Conclusion}
\label{sec:conclusion}

We developed a novel accelerated gradient-based learning algorithm, dubbed AGMA, to solve a distributed optimization problem over noisy fading MAC. We established a finite-sample bound of the error for both convex and strongly convex loss functions with Lipschitz gradient. We showed theoretically that for the strongly convex case AGMA approaches the best-known linear convergence rate as the network increases, and for the convex case AGMA improves the sub-linear convergence rate as compared to existing methods. We presented extensive simulation results using real datasets that demonstrate the better performance by AGMA as compared to existing methods.

\section{Appendix}

In this appendix we provide the proofs for Theorems \ref{th:error_bound_fading_channel} and \ref{th:error_bound_fading_channel_convex}. We start by providing lemmas and remarks that will be used throughout the proofs.

\subsection{Lemmas and Remarks}
\label{app:lemmas}

\noindent
\begin{lemma}
Let $f(\boldsymbol{x})$ be a $\mu$-strongly convex function with domain $\boldsymbol{\chi}$. Then, the following inequality holds:
\begin{equation}
\begin{array}{l}
\label{lemma:strong_convexity_inequality}
f(\boldsymbol{x}) \geq
f(\boldsymbol{y})+\left\langle\boldsymbol{\nabla}{f(\boldsymbol{y})},\boldsymbol{x}-\boldsymbol{y} \right\rangle + \frac{\mu}{2}\|\boldsymbol{x}-\boldsymbol{y}\|^{2} \hspace{0.2cm}
\forall \boldsymbol{x},\boldsymbol{y} \in \raisebox{2pt}{$\boldsymbol{\chi}$}. 
\end{array}
\end{equation}
\end{lemma}

\noindent
\begin{lemma}
\label{lemma:lip_ineuality}
Let $f(\boldsymbol{x})$ denote a convex function with domain $\boldsymbol{\chi}$, and $L$-Lipschitz gradient. Then, the following inequality holds:
\begin{equation}
\label{eq:2.1.7}
\begin{array}{l}
\displaystyle\frac{1}{2L}||\boldsymbol{\nabla} f(\boldsymbol{x}) -\boldsymbol{\nabla} f(\boldsymbol{y})||^2
\leq f(\boldsymbol{y}) - f(\boldsymbol{x}) - \left<\boldsymbol{\nabla} f(\boldsymbol{x}), \boldsymbol{y}-\boldsymbol{x} \right>
\vspace{0.2cm}\\\hspace{3.7cm}
\displaystyle\leq \frac{L}{2}||\boldsymbol{x}-\boldsymbol{y}||^2 \hspace{0.2cm}
\forall \boldsymbol{x},\boldsymbol{y} \in \raisebox{2pt}{$\boldsymbol{\chi}$}.
\end{array}
\end{equation}
\end{lemma}

The proofs for these lemmas can be found in \cite{Nesterov2004IntroductoryLO}.

Finally, note that for $f(\boldsymbol{x})$ $\mu$-strongly convex function with $L$-Lipschitz gradient and domain $\boldsymbol{\chi}$, it holds that $\mu \leq L$. This corollary can be shown by the following:
\begin{equation}
\begin{array}{l}
\mu\|\boldsymbol{x}-\boldsymbol{y}\|^2
\leq \left<\boldsymbol{\nabla} f(\boldsymbol{x}) - \boldsymbol{\nabla} f(\boldsymbol{y}), \boldsymbol{x}-\boldsymbol{y} \right>
\\\hspace{1.7cm}
\leq \|\boldsymbol{x}-\boldsymbol{y}\|\|\boldsymbol{\nabla} f(\boldsymbol{x}) - \boldsymbol{\nabla} f(\boldsymbol{y})\|
\\\hspace{1.7cm}
\leq L\|\boldsymbol{x}-\boldsymbol{y}\|^2,
\end{array}
\end{equation}
for any $\boldsymbol{x},\boldsymbol{y} \in \chi$, where the first inequality follows by \eqref{mu_lemma}, the second by Cauchy–Schwarz inequality, and the third by \eqref{L_lemma}.

\subsection{Proof of Theorem \ref{th:error_bound_fading_channel}}
\label{app:proof1}

To prove the theorem, we analyze the combined effect of the gradient-learning step and the momentum step under fading channels and additive noise. The fading channel causes distortion to the received momentum-based gradient, due to the multiplication of each local gradient by a different channel gain. The additive channel noise corrupts the convergence, which is expected to be mitigated as the SNR increases (as will be quantified analytically). To handle the effect of diverged error, we introduce new auxiliary functions that incorporate random iterated momentum-based gradient, and a new generic parameter to guard against diverged error. These analytic developments allow us to sum the error iteratively, while keeping the error in the desired convergence regime in the presence of random gradient distortion and additive noise. The tuning parameters (learning rate, and momentum series) are selected judiciously to ensure this property at each step. Finally, this allows us to upper bound the expected objective using the auxiliary functions. To the best of our knowledge, our theoretical analysis provides the first results that guarantee convergence analytically in accelerated GD over noisy fading MAC. For the ease of presentation, we divide the proof into five steps. In steps 1,2 we define the auxiliary sequence, $\Phi_k(\theta)$. Then, in steps 3,4 ,we show that it upper bounds our error term. Lastly, in step 5, we derive and establish the error bound of the theorem.

\noindent
\vspace{0.5 cm}
\emph{Step 1: Constructing a series of auxiliary functions:}\\
Define strongly convex quadratic function $\Phi_{k}(\cdot)$, $\forall k\geq0$ by induction as follow:
\begin{eqnarray}
\Phi_{0}(\boldsymbol{\theta}) &{=}& F(\boldsymbol{\theta_{0}})+\frac{\gamma_{0}}{2}\|\boldsymbol{\theta}-\boldsymbol{\theta_{0}}\|^{2} \label{phi0} \\
\Phi_{k+1}(\boldsymbol{\theta}) &{=}& (1-\alpha_{k})\Phi_{k}(\boldsymbol{\theta})  \nonumber \\&{\quad+}&  \alpha_{k}\big[F(\boldsymbol{z_{k}})+\frac{1}{\mu_{h}}\boldsymbol{v_{k}}^\intercal(\boldsymbol{\theta}-\boldsymbol{z_{k}})\nonumber \\
&{ }&\hspace{1.5cm}+\frac{\mu}{2}\|\boldsymbol{\theta}-\boldsymbol{z_{k}}\|^{2}+\epsilon_{N}\big], \label{defPhi}
\end{eqnarray}
where 
$\{\alpha_{k}\}_{k=0}^{\infty}$ is a deterministic scalar sequence that satisfies $\alpha_{k}\in (\sqrt{\frac{\mu}{\widetilde{L_\beta}}},1)$, and $\epsilon_{N}$ is some non-negative scalar. 
The use of auxiliary functions are inspired by the estimate sequence technique, first introduced by Nesterov \cite{Nesterov2004IntroductoryLO} for a centralized deterministic optimization with direct access to the true gradient. This paper is the first to develop the method for tackling a noisy distorted gradient due to the analog aggregation of the local gradients over fading MAC. For this, we design the auxiliary function such that it is computed based on the noisy estimate of the gradient $\boldsymbol{v}_{k}$. We also correct the auxiliary function by the error term $\epsilon_N$ (decreasing with $N$), which is shown later to guard against the diverged error.

In order to evaluate $\mathbb{E}[\Phi_{k+1}(\boldsymbol{\theta})]$, we first evaluate the expected value of $\boldsymbol{v_k}$ with respect to the random additive noise and channel gain processes up to time $k$:
\begin{eqnarray}
\mathbb{E}[\boldsymbol{v_k}] &{=}& 
\mathbb{E}\left [\frac{1}{N} \sum_{n=1}^N  \mu_{h}\boldsymbol{\nabla{f_{n}}}(\boldsymbol{z_k}) +\boldsymbol{w_k}\right ] \nonumber \\
&{=}& \mu_{h}\mathbb{E}[\nabla{F(\boldsymbol{z_k})}]+\mathbb{E}[\boldsymbol{w_k}] \nonumber \\
&{=}& \mu_{h}\mathbb{E}[\nabla{F(\boldsymbol{z_k})}] \label{E_vk},
\end{eqnarray}
where the second equality follows since $h_{n,k}$ is independent of $\boldsymbol{z_k}$. Note that $\boldsymbol{v}_{k}$ is a biased estimator, thus we divide it by $\mu_h$ in the auxiliary functions.
Next, define: 
\begin{eqnarray}
\lambda_{k+1} = (1-\alpha_{k})\lambda_{k}\;\; ,\;\; \lambda_{0} = 1. \label{defLambda}
\end{eqnarray}

Then, the following holds:
\begin{equation}
    \mathbb{E}[\Phi_{k}(\boldsymbol{\theta})]\leq(1-\lambda_{k})F(\boldsymbol{\theta})+\lambda_{k}\Phi_{0}(\boldsymbol{\theta}) +\epsilon_{N}. \label{est_seq_ieq}
\end{equation}
Then, (\ref{est_seq_ieq}) can be proven by induction, and by taking an expectation over \eqref{defPhi}, using (\ref{E_vk}) and \eqref{lemma:strong_convexity_inequality}. \vspace{0.3cm} 

\noindent
\emph{Step 2: Development of a canonical form:}\\
In this step we show that $\Phi_k$ is a sequence of quadratic functions, so we can exploit the simplification by the quadratic properties. For this, we use the function $\Phi_{0}(\boldsymbol{\theta}) = \Phi_{0}^{*}+\frac{\gamma_{0}}{2}\|\boldsymbol{\theta}-\boldsymbol{p_{0}}\|^{2}$ that was introduced in \cite{Nesterov2004IntroductoryLO}, and define the sequence $\{\gamma_{k}\}$ as follows:
\begin{eqnarray}
\gamma_{k+1} &{=}& (1-\alpha_{k})\gamma_{k}+\alpha_{k}\mu. \label{defGamma}
\end{eqnarray}
Note that for $\alpha_k>0\;\forall k\geq0$ we can easily get by induction that $\gamma_k > 0\; \forall k\geq0$.
Next, to prove that \{${\Phi_{k}(\boldsymbol{\theta})}$\} is a series of quadratic functions. We start by showing that $\boldsymbol{\nabla}^{2}{\Phi_{k}(\boldsymbol{\theta})} = \gamma_{k} \boldsymbol{I}_{d}$
 holds for all $k\geq0$ by induction. For the base step we have: $\boldsymbol{\nabla}^{2}{\Phi_{0}(\boldsymbol{\theta})} = \gamma_{0} \boldsymbol{I}_{d}$. For the induction step, by differentiating (\ref{defPhi}) twice we get:
\begin{eqnarray}
\boldsymbol{\nabla}^{2}{\Phi_{k+1}(\boldsymbol{\theta})} &{=}& 
(1-\alpha_{k})\boldsymbol{\nabla}^{2}{\Phi_{k}(\boldsymbol{\theta})}+ \alpha_{k}\mu \boldsymbol{I_{d}} \nonumber \\
&{=}&
\left((1-\alpha_{k})\gamma_{k}+\alpha_{k}\mu\right) \boldsymbol{I_{d}} \nonumber \\
&{=}& \gamma_{k+1} \boldsymbol{I_{d}}, \nonumber
\end{eqnarray}
where the second equality follows by the induction hypothesis. Hence $\Phi_{k+1}(\boldsymbol{\theta})$ is of the form: 
\begin{equation}
    \Phi_{k+1}(\boldsymbol{\theta}) = \Phi_{k+1}^{*} + \frac{\gamma_{k+1}}{2}\|\boldsymbol{\theta}-\boldsymbol{p_{k+1}}\|^2, \label{canonial}
\end{equation}
for some arbitrary sequences $\{\boldsymbol{p_{k}}\}$ and $\{\Phi_{k}^{*}\}$. Later, we will pick specific $p_{k}$ to achieve the desired error bound. Next, by differentiating (\ref{canonial}) we get:
\begin{equation}
    \boldsymbol{\nabla}{\Phi_{k+1}}(\boldsymbol{\theta}) = \gamma_{k+1}(\boldsymbol{\theta}-\boldsymbol{p_{k+1}}).
\end{equation}
Therefore, we get that $\Phi_{k+1}(\cdot)$ achieves extreme value at $\boldsymbol{p_{k+1}}$, i.e., $\boldsymbol{\nabla}{\Phi_{k+1}(\boldsymbol{p_{k+1}})} = 0$. Since $\gamma_{k+1} > 0$ we get that this point is minimum. As a result, substituting $\boldsymbol{p_{k+1}}$ into (\ref{canonial}) yields that $\Phi_{k+1}(\cdot)$ minimum value is $\Phi_{k+1}^{*}$. Now, by differentiating (\ref{defPhi}) we get:
\begin{equation}
    \boldsymbol{\nabla}{\Phi_{k+1}(\theta)} = (1-\alpha_{k})\gamma_{k}(\boldsymbol{\theta}-\boldsymbol{p_{k}})+\frac{\alpha_{k}\boldsymbol{v_{k}}}{\mu_{h}}+\alpha_{k}\mu(\boldsymbol{\theta-z_{k}}). 
\end{equation}
Since the minimum of $\Phi_{k+1}(\cdot)$ is obtained at $\boldsymbol{p_{k+1}}$ we get:
\begin{eqnarray}
    (1-\alpha_{k})\gamma_{k}(\boldsymbol{p_{k+1}-p_{k}})+\frac{\alpha_{k}\boldsymbol{v_{k}}}{\mu_{h}}+\alpha_{k}\mu(\boldsymbol{p_{k+1}-z_{k}}) = 0, \nonumber 
\end{eqnarray}
which yields
\begin{eqnarray}
\boldsymbol{p_{k+1}} = \frac{1}{\gamma_{k+1}}[\boldsymbol{p_{k}}\gamma_{k}(1-\alpha_{k}) + \alpha_{k}\mu \boldsymbol{z_{k}}-\frac{\alpha_{k}\boldsymbol{v_{k}}}{\mu_{h}}]. \label{defpk}
\end{eqnarray}

\noindent
\emph{Step 3: Developing $\{\Phi_{k}^{*}\}$}:

To derive $\{\Phi_{k}^{*}\}$ note that: 
\begin{eqnarray}
 &{}&
\Phi_{k+1}^{*}+\frac{\gamma_{k+1}}{2}\|\boldsymbol{z_{k}-p_{k+1}}\|^2 \nonumber \\
&\stackrel{(\ref{canonial})}{=}& 
\Phi_{k+1}(\boldsymbol{z_{k}}) \nonumber \\
&\stackrel{(\ref{defPhi})}{=}& 
(1-\alpha_{k})\Phi_{k}+
\alpha_{k}\big[F(\boldsymbol{z_{k}})+\frac{1}{\mu_{h}}\boldsymbol{v_{k}^\intercal}(\boldsymbol{z_{k}-z_{k}})\nonumber \\
&{}& \hspace{3cm} +
\frac{\mu}{2}\|\boldsymbol{z_{k}-z_{k}}\|^{2}+\epsilon_{N}\big] \nonumber \\
&{=}&
(1-\alpha_{k})(\Phi_{k}^{*}+ \frac{\gamma_{k}}{2}\|\boldsymbol{z_{k}-p_{k}}\|^2)  \nonumber \\ 
&{}& \hspace{3cm} + \alpha_{k}(F(\boldsymbol{z_{k}})+\epsilon_{N}). \label{eqPhi}
\end{eqnarray}

Using the definitions of $p_{k+1}$ in \eqref{defpk}, and $\gamma_{k+1}$ in \eqref{defGamma}, we get:
\begin{eqnarray}
 &{}&
\frac{\gamma_{k+1}}{2}\|\boldsymbol{p_{k+1}-z_{k}}\|^{2} \nonumber \\ &{=}&  \frac{1}{2\gamma_{k+1}}\Big[\|\boldsymbol{p_{k}-z_{k}}\|^{2}\gamma_{k}^{2}(1-\alpha_{k})^{2}\nonumber \\ &{\quad-}& 2\frac{\alpha_{k}(1-\alpha_{k})}{\mu_{h}}\gamma_{k}\langle v_{k},\boldsymbol{p_{k}-z_{k}}\rangle + \frac{\alpha_{k}^{2}}{\mu_{h}^{2}}\|\boldsymbol{v_{k}}\|^{2} \Big].  \label{normeq}
\end{eqnarray}

By induction, using \eqref{eqPhi}, \eqref{normeq}, and the definition of $p_{k+1}$ and $\gamma_{k+1}$, we have:
\begin{eqnarray}
\Phi_{k+1}^{*} = 
(1-\alpha_{k})\Phi_{k}^{*} + \alpha_{k}(F(\boldsymbol{z_{k}})+\epsilon_{N}) - \frac{\alpha_{k}^{2}}{2\mu_{h}^{2}\gamma_{k+1}}\|\boldsymbol{v_{k}}\|^{2} \nonumber\\\quad+ \frac{\alpha_{k}(1-\alpha_{k})\gamma_{k}}{\gamma_{k+1}\mu_{h}}\Big(\frac{\mu_{h}\mu}{2}\|\boldsymbol{z_{k}-p_{k}}\|^2   + \langle \boldsymbol{v_{k}},\boldsymbol{p_{k}-z_{k}}\rangle \Big). \label{Phi_k_star}\nonumber
\end{eqnarray}

\noindent
\emph{Step 4: Upper bounding the expected objective using the auxiliary functions:}

Next, we show by induction over $k$ that the following property holds:
\begin{eqnarray} 
\mathbb{E}[F(\boldsymbol{\theta_{k}})] &{\leq}&
\mathbb{E}[\min_{\boldsymbol{\theta}\in\mathbb{R}^{d}} \Phi_{k}(\boldsymbol{\theta})].  \label{minProp}
\end{eqnarray}
As we noticed earlier, from (\ref{canonial}) we get $\min_{\boldsymbol{\theta}\in\mathbb{R}^{d}} \Phi_{k}(\boldsymbol{\theta})=\Phi_{k}^{*}$. Hence, we will show equivalently that 
\begin{equation}
    \mathbb{E}[F(\boldsymbol{\theta_{k}})] \leq \mathbb{E}[\Phi_{k}^{*}] \label{min_prop_equiv}
\end{equation}
holds by induction. The base step follows from the definition of $\Phi_{0}(\cdot)$. Next, we assume that $\mathbb{E}[\Phi_{k}^{*}] \geq \mathbb{E}[F(\boldsymbol{\theta_{k}})] $ holds for some $k\geq 0$ and prove the property for $k+1$. Note that
\begin{eqnarray}
\mathbb{E}[\Phi_{k+1}^{*}] &{=}&
(1-\alpha_{k})\mathbb{E}[\Phi_{k}^{*}] + \alpha_{k}\mathbb{E}[F(\boldsymbol{z_{k}})] +\alpha_{k}\epsilon_{N} \nonumber\\&{\quad-}& \frac{\alpha_{k}^{2}}{2\mu_{h}^{2}\gamma_{k+1}}\mathbb{E}[\|\boldsymbol{v_{k}}\|^{2}] 
+ \frac{\alpha_{k}(1-\alpha_{k})\gamma_{k}}{\gamma_{k+1}\mu_{h}}\cdot \nonumber \\ 
&{ }&\mathbb{E}\Big[\frac{\mu_{h}\mu}{2}\|\boldsymbol{z_{k}-p_{k}}\|^2 + \langle \boldsymbol{v_{k}},\boldsymbol{p_{k}-z_{k}}\rangle \Big] \nonumber \\
&{\geq}& 
(1-\alpha_{k})\mathbb{E}[F(\boldsymbol{\theta_{k}})] + \alpha_{k}\mathbb{E}[F(\boldsymbol{z_{k}})]+\alpha_{k}\epsilon_{N} \nonumber\\&{\quad-}& \frac{\alpha_{k}^{2}}{2\mu_{h}^{2}\gamma_{k+1}}\mathbb{E}[\|\boldsymbol{v_{k}}\|^{2}] + \frac{\alpha_{k}(1-\alpha_{k})\gamma_{k}}{\gamma_{k+1}\mu_{h}} \nonumber \\ 
&{ }&\cdot\mathbb{E}\Big[\frac{\mu_{h}\mu}{2}\|\boldsymbol{z_{k}-p_{k}}\|^2 + \langle \boldsymbol{v_{k}},\boldsymbol{p_{k}-z_{k}}\rangle \Big], \nonumber 
\end{eqnarray}
where the last inequality is by the induction assumption. \\We proceed to lower bound $\mathbb{E}[\Phi_{k+1}^{*}]$ by:
\begin{eqnarray}
&{\geq}&
(1-\alpha_{k})\Big(\mathbb{E}[F(\boldsymbol{z_{k}})]  + \mathbb{E}[\langle\boldsymbol{\nabla{F(\boldsymbol{z_{k}})}},\boldsymbol{\theta_{k}-z_{k}}\rangle] \Big) \nonumber\\&{\quad+}& \alpha_{k}\mathbb{E}[F(\boldsymbol{z_{k}})] +\alpha_{k}\epsilon_{N} - \frac{\alpha_{k}^{2}}{2\mu_{h}^{2}\gamma_{k+1}}\mathbb{E}[\|\boldsymbol{v_{k}}\|^{2}] 
\nonumber\\&{\quad+}&  \frac{\alpha_{k}(1-\alpha_{k})\gamma_{k}}{\gamma_{k+1}\mu_{h}} \nonumber \\ 
&{ }&\cdot\mathbb{E}\Big[\frac{\mu_{h}\mu}{2}\|\boldsymbol{z_{k}-p_{k}}\|^2 + \langle \mu_{h}\boldsymbol{\nabla{F(\boldsymbol{z_{k}})}},\boldsymbol{p_{k}-z_{k}}\rangle \Big]  \nonumber \\
&{=}&
\mathbb{E}[F(\boldsymbol{z_{k}})]-\frac{\alpha_{k}^{2}}{2\mu_{h}^{2}\gamma_{k+1}}\mathbb{E}[\|\boldsymbol{v_{k}}\|^{2}] +\alpha_{k}\epsilon_{N}\nonumber\\&{\quad-}& \frac{\alpha_{k}(1-\alpha_{k})\gamma_{k}}{\gamma_{k+1}}\frac{\mu}{2}\|\boldsymbol{z_{k}-p_{k}}\|^2
\nonumber\\&{\quad+}&
(1-\alpha_{k})\mathbb{E}\big[\langle\boldsymbol{\nabla{F(z_{k})}},\boldsymbol{\theta_{k}-z_{k}}\nonumber\\
&{ }&\hspace{2cm}+\frac{\alpha_{k}\gamma_{k}}{\gamma_{k+1}}(p_{k}-z_{k})\rangle\big],
 \label{ineqPhi} 
\end{eqnarray} 
where the second inequality is due to convexity of $F(\cdot)$.

To proceed, we next develop the term $\mathbb{E}[\|\boldsymbol{v_k}\|^2]$:
\begin{eqnarray}
\mathbb{E}[\|\boldsymbol{v_k}\|^2] &{=}& 
\mathbb{E}\left [\left\|\frac{1}{N} \sum_{n=1}^N h_{n,k}\boldsymbol{\nabla{f_{n}}}(\boldsymbol{z_k}) +\boldsymbol{w_k} \right\|^{2} \right ] \nonumber \\ 
&{=}&
\mathbb{E}\left [\Bigg\|\frac{1}{N} \sum_{n=1}^N h_{n,k}\boldsymbol{\nabla{f_{n}}}(\boldsymbol{z_k}) \Bigg\|^{2} \right ] \nonumber \\
&{\quad+}& 
2\mathbb{E}\left[\left(\frac{1}{N} \sum_{n=1}^N h_{n,k}\boldsymbol{\nabla{f_{n}}}(\boldsymbol{z_k})^\intercal\boldsymbol{w_k}\right)\right] \nonumber \\
&{\quad+}&
\mathbb{E}\left[\left|\left|\boldsymbol{w_k}\right|\right|^2 \right] \nonumber
\end{eqnarray}
\begin{eqnarray}
&{=}&
\frac{1}{N^2}\sum_{n,m =1}^N \mathbb{E}\Big[(h_{n,k} \boldsymbol{\nabla{f_{n}}}(\boldsymbol{z_k}))^\intercal (h_{m,k}  \boldsymbol{\nabla{f_{m}}}(\boldsymbol{z_k}))\Big] \nonumber \\ &{\quad+}&
\frac{ d\sigma_w^2} {E_N N^2}\nonumber \\
&{=}&
\mu_{h}^{2}\mathbb{E}[\|\boldsymbol{\nabla{F(\boldsymbol{z_{k}})}}\|^{2}]+\frac{\sigma_{h}^{2}}{N^{2}}\sum_{n=1}^{N} \mathbb{E}[\|\boldsymbol{\nabla{f_{n}(\boldsymbol{z_{k}})}}\|^{2}] \nonumber \\ 
&{\quad+}&
\frac{d\sigma_{w}^{2}}{E_{N}N^{2}}. \label{var_vk}
\end{eqnarray}

Inequality (\ref{eq:2.1.7}) yields:
\begin{eqnarray}
\hspace{-0.3cm}F(\boldsymbol{\theta_{k+1}}) -F(\boldsymbol{z_{k}})-\langle\boldsymbol{\nabla{F(\boldsymbol{z_{k}})}},\boldsymbol{\theta_{k+1}}-\boldsymbol{z_{k}}\rangle \leq \frac{L}{2}\|\boldsymbol{z_{k}}-\boldsymbol{\theta_{k+1}}\|^{2}, \nonumber 
\end{eqnarray}
and substituting (\ref{eq:theta_k_1}) yields: \begin{eqnarray}
F(\boldsymbol{\theta_{k+1}}) -F(\boldsymbol{z_{k}})-\langle\boldsymbol{\nabla{F(\boldsymbol{z_{k}})}},-\beta \boldsymbol{v_{k}}\rangle \leq \frac{L}{2}\|\boldsymbol{\beta v_{k}}\|^{2}. \nonumber 
\end{eqnarray}
Taking expectation of both sides of the equation and using (\ref{var_vk}) yields:
\begin{eqnarray}
\mathbb{E}[F(\boldsymbol{\theta_{k+1}})] &{-}& \mathbb{E}[F(\boldsymbol{z_{k}})] + \frac{\beta}{\mu_{h}}\mathbb{E}[\|\boldsymbol{v_{k}}\|^{2}] \nonumber \\ 
&\hspace{-1cm}-&\hspace{-0.5cm} \frac{\beta}{\mu_{h}}\left(\frac{\sigma_{h}^{2}G(N)}{N}+\frac{d\sigma_{w}^{2}}{E_{N}N^{2}}\right) 
\leq\frac{L\beta^{2}}{2}\mathbb{E}[\|\boldsymbol{v_{k}}\|^{2}]\nonumber.  
\end{eqnarray}

Then, we get:
\begin{eqnarray}
\label{eq:th1_Ef_theta_k_1}
\mathbb{E}[F(\boldsymbol{\theta_{k+1}})] &{\leq}& \mathbb{E}[F(\boldsymbol{z_{k}})]-\frac{\beta}{2}\Big(\frac{2}{\mu_{h}}-\beta L\Big)\mathbb{E}[\|\boldsymbol{v_{k}}\|^{2}] \nonumber \\ &{+}& \delta_{N},
\end{eqnarray}
where we define 
\begin{equation}
\delta_{N} \triangleq \frac{\beta}{\mu_{h}}\left(\frac{\sigma_{h}^{2}G(N)}{N}+\frac{d\sigma_{w}^{2}}{E_{N}N^{2}}\right).
\end{equation}
Let us choose $\{\alpha_{k}\}$ as follows:
\begin{eqnarray}
    \alpha_{k}^2 &{=}&  \frac{1}{\widetilde{L_\beta}}\gamma_{k+1}, \label{defAlpha}
\end{eqnarray}
where $\widetilde{L_\beta}$ is defined in \eqref{th:L_beta}. Since we require positive $\widetilde{L_\beta}$ we derive the range of the step size, where the algorithm convergence is guaranteed:
 \begin{equation}
     \beta\Big(\frac{2}{\mu_{h}}-\beta L\Big)\mu_h^2 > 0 \quad \Rightarrow \quad 0<\beta<\frac{2}{\mu_h L}. \label{beta_bound}
 \end{equation}
Also note:
 \begin{eqnarray}
    \frac{1}{L}-\frac{1}{\widetilde{L_\beta}} \quad =
    & L\mu_h^2\big(\beta-\frac{1}{L\mu_h}\big)^2 \geq 0 \nonumber\\
    \Rightarrow& \widetilde{L_\beta} \geq L .
 \end{eqnarray}
Combined with (\ref{defGamma}) we get
\begin{equation}
\alpha_{k+1}^{2} = (1-\alpha_{k+1})\alpha_{k}^{2}+\frac{\mu}{\widetilde{L_\beta}}\alpha_{k+1}. \nonumber
\end{equation}

In Step 5 we ensure that our choice of $\alpha_{k}$ is valid. 
Next, using (\ref{defAlpha}) yields:
\begin{eqnarray}
\frac{\alpha_{k}^{2}}{2\mu_{h}^{2}\gamma_{k+1}} = \frac{1}{2\mu_{h}^{2}\widetilde{L_\beta}}. \nonumber
\end{eqnarray} 
Therefore, 
\begin{eqnarray}
\mathbb{E}[F(\boldsymbol{\theta_{k+1}})]-\delta_{N}\leq
\mathbb{E}[F(\boldsymbol{z_{k}})]-\frac{\alpha_{k}^{2}}{2\mu_{h}^{2}\gamma_{k+1}}\mathbb{E}[\|\boldsymbol{v_{k}}\|^{2}]. \label{step4.1}
\end{eqnarray}

Next, substituting (\ref{step4.1}) into (\ref{ineqPhi}) yields:
\begin{eqnarray}
\mathbb{E}[\Phi_{k+1}^{*}] &{\geq}& 
\mathbb{E}[F(\boldsymbol{\theta_{k+1}})]+ (\alpha_{k}\epsilon_{N}-\delta_{N}) \nonumber\\&{\quad-}&
\frac{\alpha_{k}(1-\alpha_{k})\gamma_{k}}{\gamma_{k+1}}\frac{\mu}{2}\|\boldsymbol{z_{k}}-\boldsymbol{p_{k}}\|^2
\nonumber\\&{\quad+}&
(1-\alpha_{k})\mathbb{E}\Big[\langle\boldsymbol{\nabla{F(\boldsymbol{z_{k}})}},\boldsymbol{\theta_{k}}-\boldsymbol{z_{k}} \nonumber\\ &{ }& \hspace{2cm} +\frac{\alpha_{k}\gamma_{k}}{\gamma_{k+1}}(\boldsymbol{p_{k}}-\boldsymbol{z_{k}})\rangle\Big]. \label{induction_step}
\end{eqnarray}
Note that $\{\boldsymbol{z_{k}}\}$ is an arbitrary sequence (which we will show would be the momentum-based update sequence). Hence, we can choose it as follows to cancel the rightmost term:
\begin{eqnarray}
\boldsymbol{\theta_{k}}-\boldsymbol{z_{k}}+\frac{\alpha_{k}\gamma_{k}}{\gamma_{k+1}}(\boldsymbol{p_{k}}-\boldsymbol{z_{k}}) = 0,
\end{eqnarray}
and combining with (\ref{defGamma}) yields: 
\begin{eqnarray}
\boldsymbol{z_{k}} &{=}&
\frac{\alpha_{k}\gamma_{k}\boldsymbol{p_{k}}+\boldsymbol{\theta_{k}}\gamma_{k+1}}{\alpha_{k}\mu+\gamma_{k}}. \label{defZk}
\end{eqnarray}

Next, we derive lower bound for $\alpha_{k}$. Recall the initialization condition of $\alpha_{0}$ is $\alpha_{0} > \sqrt{\frac{\mu}{L}}$. Thus,
\begin{eqnarray}
\gamma_{0} =  \frac{\alpha_{0}(\alpha_{0}L-\mu)}{1-\alpha_{0}} >
\frac{\sqrt{\frac{\mu}{L}}(\sqrt{\frac{\mu}{L}}L-\mu)}{1-\sqrt{\frac{\mu}{L}}}
=
\frac{\mu(1-\sqrt{\frac{\mu}{L}})}{1-\sqrt{\frac{\mu}{L}}}
=
\mu, \nonumber
\end{eqnarray}
Assume $\gamma_{k}>\mu$. Then, by induction:
\begin{eqnarray}
\gamma_{k+1} &\stackrel{(\ref{defGamma})}{=}&
(1-\alpha_{k})\gamma_{k}+\alpha_{k}\mu 
>(1-\alpha_{k})\mu+\alpha_{k}\mu
=\mu.  \nonumber
\end{eqnarray}
Therefore,
\begin{eqnarray}
\alpha_{k}  &\stackrel{(\ref{defAlpha})}{=}&
\sqrt{\frac{\gamma_{k+1}}{\widetilde{L_\beta}}} > \sqrt{\frac{\mu}{\widetilde{L_\beta}}}\;\;\;\; \forall k\geq 0. \label{alpha_lower_bound}
\end{eqnarray}
Let the error term be:
\begin{equation}
    \epsilon_{N} = \sqrt{\frac{\widetilde{L_\beta}}{\mu}}\delta_{N} = 
    \sqrt{\frac{\widetilde{L_\beta}}{\mu}}\frac{\beta}{\mu_{h}}\left(\frac{\sigma_{h}^{2}G(N)}{N}+\frac{d\sigma_{w}^{2}}{E_{N}N^{2}}\right).\label{epsilon_n_def}
\end{equation}
Thus, we get $(\alpha_{k}\epsilon_{N}-\delta_{N}) \geq 0$. Furthermore, note that
\begin{equation}
    -\frac{\alpha_{k}(1-\alpha_{k})\gamma_{k}}{\gamma_{k+1}}\frac{\mu}{2} \geq 0 \label{observation_1}
\end{equation}
holds.

Finally, combining (\ref{defZk}) , (\ref{epsilon_n_def}) , (\ref{observation_1}) on (\ref{induction_step})  yields $\mathbb{E}[F(\boldsymbol{\theta_{k+1}})] \leq \mathbb{E}[\Phi_{k+1}^{*}]$, which shows that (\ref{minProp}) holds $\forall k\geq0$.\vspace{0.2cm}

\noindent
\emph{Step 5: Establishing the error bound:}

We start by substituting (\ref{defZk}) into (\ref{defpk}) and rearranging terms to get:    
\begin{equation}
   \boldsymbol{p_{k+1}} = \boldsymbol{\theta_{k}}+\frac{1}{\alpha_{k}}\Big(\boldsymbol{\theta_{k+1}}-\boldsymbol{\theta_{k}}\Big). \nonumber
\end{equation}
Substituting the above into (\ref{defZk}) and using (\ref{defGamma}) yield:
\begin{equation}
\boldsymbol{z_{k+1}} = \boldsymbol{\theta_{k+1}}+\eta_{k}(\boldsymbol{\theta_{k+1}}-\boldsymbol{\theta_{k}}), \nonumber
\end{equation}
where
\begin{equation}
\eta_{k} \triangleq
\frac{\alpha_{k}(1-\alpha_{k})}{\alpha_{k+1}+\alpha_{k}^{2}} \nonumber 
\end{equation}
is simplified using (\ref{defAlpha}).\vspace{0.2cm}

Before completing the error bound, we need to show that the chosen series \{$\alpha_k$\} is valid. Until (\ref{defAlpha}), all we assumed is that $\alpha_k \in (0,1)\;\forall k \geq 0$. Thus, it remains to show that $\alpha_{k}^2 =  \frac{1}{\widetilde{L_\beta}}\gamma_{k+1}$ satisfies this requirement. Note that proving the following will satisfy:
\begin{equation}\label{alpha_k_induction}
    \sqrt{\frac{\mu}{\widetilde{L_\beta}}} < \alpha_k < 1,\quad \forall \mu \in[0, L),\quad \forall k\geq 0 .
\end{equation}
The left inequality is presented in (\ref{alpha_lower_bound}), which holds for the initialization condition and the assumption of $\alpha_k<1$. Note that for $\mu$-strongly convex function with $L$-Lipschitz gradient $\mu \leq L$ holds, as shown in Appendix \ref{app:lemmas}. It remains to show that $\alpha_k<1,$ $\forall k \geq 0$. We prove this by induction over $k$. The base step follows by the initialization condition: $\alpha_0<1$. We assume for some $k\geq0$ that $\alpha_k<1$ holds and prove it for $k+1$. For the induction step by (\ref{defAlpha}) we have: 
 \begin{eqnarray}
     \alpha_{k+1}^{2} + (\alpha_{k}^{2}-\frac{\mu}{\widetilde{L_\beta}})\alpha_{k+1}-\alpha_{k}^{2}=0,
\end{eqnarray}     
and thus
\begin{eqnarray}     
     \alpha_{k+1} = \frac{-(\alpha_{k}^{2}-\frac{\mu}{\widetilde{L_\beta}})\pm \sqrt{(\alpha_{k}^{2}-\frac{\mu}{\widetilde{L_\beta}})^2+4\alpha_k^2}}{2}.
 \end{eqnarray}
 Since we know that \{$\alpha_k\} > \frac{\mu}{\widetilde{L_\beta}} \geq 0 $ is the non-negative possible solution. Also, the following holds:
 \begin{eqnarray}
     \alpha_{k+1}\quad =& \frac{-(\alpha_{k}^{2}-\frac{\mu}{\widetilde{L_\beta}})+ \sqrt{(\alpha_{k}^{2}-\frac{\mu}{\widetilde{L_\beta}})^2+4\alpha_k^2}}{2}\nonumber\\=&\frac{-(\alpha_{k}^{2}-\frac{\mu}{\widetilde{L_\beta}})+ \sqrt{(\alpha_{k}^{2}-\frac{\mu}{\widetilde{L_\beta}}+2\alpha_k)^2-4\alpha_k(\alpha_{k}^{2}-\frac{\mu}{\widetilde{L_\beta}})}}{2}.\nonumber
 \end{eqnarray}
 Since $4\alpha_k(\alpha_{k}^{2}-\frac{\mu}{\widetilde{L_\beta}}) > 0$ we get: 
 \begin{eqnarray}
    <& \frac{-(\alpha_{k}^{2}-\frac{\mu}{\widetilde{L_\beta}})+ \sqrt{(\alpha_{k}^{2}-\frac{\mu}{\widetilde{L_\beta}}+2\alpha_k)^2}}{2}  \nonumber\\=& \frac{-(\alpha_{k}^{2}-\frac{\mu}{\widetilde{L_\beta}})+ (\alpha_{k}^{2}-\frac{\mu}{\widetilde{L_\beta}}+2\alpha_k)}{2}=\alpha_k<1, 
 \end{eqnarray}
where the last inequality is the induction assumption.

Finally, we use the results of the previous steps to achieve the desired error bound:
\begin{eqnarray}
 &{}&
\mathbb{E}[F(\boldsymbol{\theta_{k}})]-F(\boldsymbol{\theta^{*}}) \nonumber \\
&\stackrel{(\ref{minProp})}{\leq}& 
\mathbb{E}\big[\min_{\boldsymbol{\theta}\in\mathbb{R}^{d}} \Phi_{k}(\boldsymbol{\theta})\big] -F(\boldsymbol{\theta^{*}}) \nonumber \\
&\stackrel{(\ref{est_seq_ieq})}{\leq}&
\mathbb{E}\Big[\min_{\boldsymbol{\theta}\in\mathbb{R}^{d}}\big\{(1-\lambda_{k})F(\boldsymbol{\theta})+
\lambda_{k}\Phi_{0}(\boldsymbol{\theta})+ \epsilon_{N} \big\}\Big]-F(\boldsymbol{\theta^{*}}) \nonumber \\
&{\leq}&
(1-\lambda_{k})F(\boldsymbol{\theta^{*}}) +
\lambda_{k}\Phi_{0}(\boldsymbol{\theta^{*}})-F(\boldsymbol{\theta^{*}})+\epsilon_{N} \nonumber \\
&{=}&
\lambda_{k}\left(\Phi_{0}(\boldsymbol{\theta^{*}})-F(\boldsymbol{\theta^{*}})\right)+\epsilon_{N} \nonumber \\ 
&\stackrel{(\ref{phi0})}{=}&
\lambda_{k}\Big(F(\boldsymbol{\theta_{0}})-F(\boldsymbol{\theta^{*}})
+\frac{\gamma_{0}}{2}\|\boldsymbol{\theta_{0}}-\boldsymbol{\theta^{*}}\|^{2}\Big) + \epsilon_{N}. \label{bound}
\end{eqnarray} 
Now, from (\ref{defLambda}) and (\ref{alpha_lower_bound}) we get:
\begin{eqnarray}
\label{eq:condition_number}
\lambda_{k} = \prod_{i=0}^{k-1} (1-\alpha_{i}) \leq
\Big(1-\sqrt{\frac{\mu}{\widetilde{L_\beta}}}\Big)^{k}. \label{lambda_inq1}
\end{eqnarray}
Combining (\ref{lambda_inq1}) and (\ref{bound}) yields:
\begin{equation}
\begin{array}{l}
\displaystyle
\mathbb{E}[F(\boldsymbol{\theta_k})]-F(\boldsymbol{\theta^*}) \leq \vspace{0.2cm}\nonumber\\ \Big(1-\sqrt{\frac{\mu}{\widetilde{L_\beta}}}\Big)^{k}\Big(F(\boldsymbol{\theta_{0}})-F(\boldsymbol{\theta^{*}})
+\frac{\gamma_{0}}{2}\|\boldsymbol{\theta_{0}}-\boldsymbol{\theta^{*}}\|^{2}\Big) + \epsilon_{N}, \\\hspace{0.3cm}
\end{array}
\end{equation}
which competes the proof.
$\qed$

\subsection{Proof of Theorem \ref{th:error_bound_fading_channel_convex}}
\label{app:proof2}  

Due to space limitations, we focus on the main changes required to proving Theorem \ref{th:error_bound_fading_channel_convex}, without reproducing steps which are similar to the proof of Theorem \ref{th:error_bound_fading_channel}.

\noindent
\emph{Step 1: Constructing a series of auxiliary functions:}

We construct a series of auxiliary functions as in \eqref{phi0}, \eqref{defPhi}, while setting $\mu=0$ (since the objective function is only convex), and $\epsilon_{N}=0$. $\lambda_k$ is defined as in \eqref{defLambda}.

\noindent
\emph{Steps 2 and 3: Development of a canonical form and developing $\{\Phi_{k}^{*}\}$:}

These steps are similar to the proof of Theorem \ref{th:error_bound_fading_channel}, while setting $\mu=0$, and $\epsilon_{N}=0$. We get:

\begin{eqnarray}
\Phi_{k+1}^{*} &=&
(1-\alpha_{k})\Phi_{k}^{*} + \alpha_{k}F(\boldsymbol{z_{k}})  + \frac{\alpha_{k}}{\mu_{h}}\langle \boldsymbol{v_{k}},\boldsymbol{p_{k}}-\boldsymbol{z_{k}}\rangle  \nonumber \\&{\quad-}& \frac{\alpha_{k}^{2}}{2\mu_{h}^{2}\gamma_{k+1}}\|\boldsymbol{v_{k}}\|^{2}, \nonumber
\end{eqnarray}
and 
\begin{eqnarray}
\boldsymbol{p_{k+1}} = \frac{1}{\gamma_{k+1}}[\boldsymbol{p_{k}}\gamma_{k}(1-\alpha_{k}) -\frac{\alpha_{k}\boldsymbol{v_{k}}}{\mu_{h}}]. \label{defpk_convex}
\end{eqnarray}

\noindent
\emph{Step 4: Upper bounding the expected objective using the auxiliary functions:}

In this step the bound on $\mathbb{E}[F(\boldsymbol{\theta_{k}})]$ need to be shown is different. Specifically, we show by induction over $k$ that the following property holds:
\begin{eqnarray} 
\mathbb{E}[F(\boldsymbol{\theta_{k}})] &{\leq}&
\mathbb{E}[\min_{\boldsymbol{\theta}\in\mathbb{R}^{d}} \Phi_{k}(\boldsymbol{\theta})] + k\delta_{N},  \label{minProp_convex}
\end{eqnarray}
where 
\begin{eqnarray}
  \delta_{N} = \frac{\beta}{\mu_{h}}\left(\frac{\sigma_{h}^{2}G(N)}{N}+\frac{d\sigma_{w}^{2}}{E_{N}N^{2}}\right). \label{delta_N_convex}
\end{eqnarray}
Since $\min_{\boldsymbol{\theta}\in\mathbb{R}^{d}} \Phi_{k}(\boldsymbol{\theta})=\Phi_{k}^{*}$, we will show equivalently that 
\begin{equation}
    \mathbb{E}[F(\boldsymbol{\theta_{k}})] \leq \mathbb{E}[\Phi_{k}^{*}] +k\delta_{N} \label{min_prop_equiv}
\end{equation}
holds by induction. The base step follows from the definition of $\Phi_{0}(\cdot)$. Next, we assume that $\mathbb{E}[\Phi_{k}^{*}]+k\delta_{N}\geq \mathbb{E}[F(\boldsymbol{\theta_{k}})]$ holds for some $k\geq 0$ and prove the property for $k+1$. Note that
\begin{eqnarray}
\mathbb{E}[\Phi_{k+1}^{*}] &{=}&
(1-\alpha_{k})\mathbb{E}[\Phi_{k}^{*}] + \alpha_{k}\mathbb{E}[F(\boldsymbol{z_{k}})] \nonumber\\&{\quad-}& \frac{\alpha_{k}^{2}}{2\mu_{h}^{2}\gamma_{k+1}}\mathbb{E}[\|\boldsymbol{v_{k}}\|^{2}] 
 + \frac{\alpha_{k}}{\mu_{h}}\mathbb{E}\Big[\langle \boldsymbol{v_{k}},\boldsymbol{p_{k}}-\boldsymbol{z_{k}}\rangle \Big] \nonumber \\
&{\geq}& 
(1-\alpha_{k})(\mathbb{E}[F(\boldsymbol{\theta_{k}})]-k\delta_{N}) + \alpha_{k}\mathbb{E}[F(\boldsymbol{z_{k}})]\nonumber\\&{\quad-}& \frac{\alpha_{k}^{2}}{2\mu_{h}^{2}\gamma_{k+1}}\mathbb{E}[\|\boldsymbol{v_{k}}\|^{2}]+ \frac{\alpha_{k}}{\mu_{h}}\mathbb{E}\Big[\langle \boldsymbol{v_{k}},\boldsymbol{p_{k}}-\boldsymbol{z_{k}}\rangle \Big], \nonumber 
\end{eqnarray}
where the last inequality is by the induction assumption. Due to convexity of $F(\cdot)$, we proceed to lower bound $\mathbb{E}[\Phi_{k+1}^{*}]$ by:
\begin{eqnarray}
\mathbb{E}[\Phi_{k+1}^{*}] &{\geq}&
\mathbb{E}[F(\boldsymbol{z_{k}})]-\frac{\alpha_{k}^{2}}{2\mu_{h}^{2}\gamma_{k+1}}\mathbb{E}[\|\boldsymbol{v_{k}}\|^{2}] -k\delta_{N}
\nonumber\\&{\hspace{-2cm}\quad+}&
\hspace{-1cm}
(1-\alpha_{k})\mathbb{E}\big[\langle\boldsymbol{\nabla{F(\boldsymbol{z_{k}})}},\boldsymbol{\theta_{k}}-\boldsymbol{z_{k}}+\frac{\alpha_{k}\gamma_{k}}{\gamma_{k+1}}(\boldsymbol{p_{k}}-\boldsymbol{z_{k}})\rangle\big]. \nonumber \\
 \label{ineqPhi_convex} 
\end{eqnarray} 

Next, we apply the same developments as in \eqref{eq:th1_Ef_theta_k_1}-\eqref{step4.1}, and combine \eqref{step4.1} with \eqref{ineqPhi_convex} to get:

\begin{eqnarray}
\mathbb{E}[\Phi_{k+1}^{*}] &{\geq}& 
\mathbb{E}[F(\boldsymbol{\theta_{k+1}})]- (k\delta_{N}+\delta_{N}) \nonumber\\&{\quad+}&
(1-\alpha_{k})\mathbb{E}\Big[\langle\boldsymbol{\nabla{F(\boldsymbol{z_{k}})}},\boldsymbol{\theta_{k}}-\boldsymbol{z_{k}} \nonumber\\ &{ }& \hspace{2cm} +\frac{\alpha_{k}\gamma_{k}}{\gamma_{k+1}}(\boldsymbol{p_{k}}-\boldsymbol{z_{k}})\rangle\Big]. \nonumber
\end{eqnarray}
Choosing $\{\boldsymbol{z_{k}}\}$ as a momentum-based update sequence in this case as well yields (\ref{minProp_convex}).

\emph{Step 5: Establishing the error bound:}

Now, we bound the error by:
\begin{eqnarray}
 &{}&
\mathbb{E}[F(\boldsymbol{\theta_{k}})]-F(\boldsymbol{\theta^{*}}) \nonumber \\
&\stackrel{(\ref{minProp_convex})}{\leq}& 
\mathbb{E}\big[\min_{\boldsymbol{\theta}\in\mathbb{R}^{d}} \Phi_{k}(\boldsymbol{\theta})\big]+k\delta_{N} -F(\boldsymbol{\theta^{*}}) \nonumber \\
&
\leq& 
\mathbb{E}\Big[\min_{\boldsymbol{\theta}\in\mathbb{R}^{d}}\big\{(1-\lambda_{k})F(\boldsymbol{\theta})+
\lambda_{k}\Phi_{0}(\boldsymbol{\theta}) \big\}\Big] +k\delta_{N}-F(\boldsymbol{\theta^{*}}) \nonumber \\
&{\leq}&
(1-\lambda_{k})F(\boldsymbol{\theta^{*}}) +
\lambda_{k}\Phi_{0}(\boldsymbol{\theta^{*}})-F(\boldsymbol{\theta^{*}})+k\delta_{N} \nonumber \\
&{=}&
\lambda_{k}\left(\Phi_{0}(\boldsymbol{\theta^{*}})-F(\boldsymbol{\theta^{*}})\right)+k\delta_{N} \nonumber \\ 
&
=&
\lambda_{k}\Big(F(\boldsymbol{\theta_{0}})-F(\boldsymbol{\theta^{*}})
+\frac{\gamma_{0}}{2}\| \boldsymbol{\theta_{0}}-\boldsymbol{\theta^{*}}\|^{2}\Big) +k\delta_{N}. \label{bound_convex}
\end{eqnarray} 

Next, we show that $\gamma_{k} = \gamma_{0}\lambda_{k}$ by induction over $k$. The base step yields: $\gamma_{0} = \gamma_{0}\lambda_{0}$. Next, we assume that $\gamma_{k}=\gamma_{0}\lambda_{k}$ and prove the induction hypothesis for $k+1$.
\begin{eqnarray}
\gamma_{k+1}  &\stackrel{(\ref{defGamma})}{=}& (1-\alpha_{k})\gamma_{k} \nonumber \\ &{=}&  (1-\alpha_{k})\lambda_{k}\gamma_{0} \stackrel{(\ref{defLambda})}{=} \gamma_{0}\lambda_{k+1}, \label{gamma_induction_convex}
\end{eqnarray}
where the second equality follows by the induction hypothesis.
Thus,
\begin{equation}
    \alpha_{k} \stackrel{(\ref{defAlpha})}{=} \sqrt{\frac{\gamma_{k+1}}{\widetilde{L_\beta}}} = \sqrt{\frac{\gamma_{0}\lambda_{k+1}}{\widetilde{L_\beta}}}. \label{ak_ieq_convex}
\end{equation}
Since $\{\lambda_{k}\}$ is a non-increasing sequence we have:
\begin{eqnarray}
\frac{1}{\sqrt{\lambda_{k+1}}} - \frac{1}{\sqrt{\lambda_{k}}} &{=}& \frac{\sqrt{\lambda_{k}} \nonumber-\sqrt{\lambda_{k+1}}}{\sqrt{\lambda_{k}}\sqrt{\lambda_{k+1}}} 
 \nonumber \\ &{=}&
\frac{\lambda_{k}-\lambda_{k+1}}{\sqrt{\lambda_{k}}\sqrt{\lambda_{k+1}}(\sqrt{\lambda_{k}}+\sqrt{\lambda_{k+1}})} \nonumber \\
&{\geq}&
\frac{\lambda_{k}-\lambda_{k+1}}{2\lambda_{k}\sqrt{\lambda_{k+1}}} 
 \stackrel{(\ref{defLambda})}{=}
\frac{\lambda_{k}-(1-\alpha_{k})\lambda_{k}}{2\lambda_{k}\sqrt{\lambda_{k+1}}} 
 \nonumber \\ &{=}&
\frac{\alpha_{k}}{2\sqrt{\lambda_{k+1}}} 
\stackrel{(\ref{ak_ieq_convex})}{=}
\frac{1}{2}\sqrt{\frac{\gamma_{0}}{\widetilde{L_\beta}}}. \nonumber
\end{eqnarray}
Therefore, we get:
\begin{eqnarray}
\frac{1}{\sqrt{\lambda_{k}}} &{\geq}& \frac{1}{\sqrt{\lambda_{k-1}}} + \frac{1}{2}\sqrt{\frac{\gamma_{0}}{\widetilde{L_\beta}}} \geq \frac{1}{\sqrt{\lambda_{k-2}}} + \frac{2}{2}\sqrt{\frac{\gamma_{0}}{\widetilde{L_\beta}}} \nonumber \\ &{\geq}& \dots \geq \frac{1}{\sqrt{\lambda_{0}}} + \frac{k}{2}\sqrt{\frac{\gamma_{0}}{\widetilde{L_\beta}}}  \stackrel{(\ref{defGamma})}{=} 1 + \frac{k}{2}\sqrt{\frac{\gamma_{0}}{\widetilde{L_\beta}}}. \nonumber 
\end{eqnarray}
Rearranging terms yields:
\begin{eqnarray}
\lambda_{k} &{\leq}& \frac{4\widetilde{L_\beta}}{\big(2\sqrt{\widetilde{L_\beta}}+k\sqrt{\gamma_{0}}\big)^2}. \label{lambda_inq2_convex}
\end{eqnarray}
Combining (\ref{bound_convex}) and (\ref{lambda_inq2_convex}) yields:
\begin{equation}
\label{eq:th_error_f_equal_convex}
\begin{array}{l}
\displaystyle 
\mathbb{E}[F(\boldsymbol{\theta_k})]-F(\boldsymbol{\theta^*}) \vspace{0.2cm}\\\hspace{0.3cm}
\displaystyle \leq  \frac{4\widetilde{L_\beta}}{\left(2\sqrt{\widetilde{L_\beta}}+k\sqrt{\gamma_{0}}\right)^{2}}(F(\boldsymbol{\theta_{0}})-F(\boldsymbol{\theta^{*}})+\frac{\gamma_{0}}{2}\|\boldsymbol{\theta_{0}}-\boldsymbol{\theta^{*}}\|^{2}) \\\hspace{7cm}+k\delta_{N}.\hspace{0.3cm} \nonumber
\end{array}
\end{equation}
Finally, setting $k_0=\lfloor N^{1-\epsilon}\rfloor$ proves the desired error bound \eqref{eq:th_error_f_equal_convex1} for all $k=1, 2, ..., k_0$. 
$\qed$

\bibliographystyle{IEEEtran}
\bibliography{bibliography}

\end{document}